\documentclass[letterpaper]{article} % DO NOT CHANGE THIS
\usepackage{aaai2026}  % DO NOT CHANGE THIS
\usepackage{times}  % DO NOT CHANGE THIS
\usepackage{helvet}  % DO NOT CHANGE THIS
\usepackage{courier}  % DO NOT CHANGE THIS
\usepackage[hyphens]{url}  % DO NOT CHANGE THIS
\usepackage{graphicx} % DO NOT CHANGE THIS
\usepackage{natbib}  % DO NOT CHANGE THIS AND DO NOT ADD ANY OPTIONS TO IT
\usepackage{caption} % DO NOT CHANGE THIS AND DO NOT ADD ANY OPTIONS TO IT
 \usepackage{amssymb}
\usepackage{algorithm}
\usepackage{algorithmic}
\usepackage{amsmath}
\defcitealias{marvel}{Sinha and Khan et al. 2025}
\usepackage{newfloat}
\usepackage{listings}
\DeclareCaptionStyle{ruled}{labelfont=normalfont,labelsep=colon,strut=off} % DO NOT CHANGE THIS
\floatstyle{ruled}
\newfloat{listing}{tb}{lst}{}
\floatname{listing}{Listing}
\title{NURBGen: High-Fidelity Text-to-CAD Generation through LLM-Driven NURBS Modeling}

\author{
    Muhammad Usama\textsuperscript{\rm 1,2,3}\textsuperscript{\rm *},
    Mohammad Sadil Khan\textsuperscript{\rm 1,2,3}\textsuperscript{\rm *}\textsuperscript{$\dagger$},
    Didier Stricker\textsuperscript{\rm 1,2},
    Muhammad Zeshan Afzal\textsuperscript{\rm 1,3}
}
\affiliations{
    \textsuperscript{\rm 1}German Research Center for Artificial Intelligence (DFKI), Kaiserslautern, Germany \\
    \textsuperscript{\rm 2}RPTU Kaiserslautern-Landau (RPTU), Germany \\
    \textsuperscript{\rm 3}MindGarage, Kaiserslautern, Germany \\
    \{muhammad.usama, mohammad.khan, didier.stricker, muhammad\_zeshan.afzal\}@dfki.de \\ [0.1cm]
    {\rm *}Equally contributing first authors. \, \textsuperscript{$\dagger$}Corresponding Author
    
}

\usepackage{bibentry}
\begin{document}
\setcounter{footnote}{1}
\maketitle

% \begingroup
% \setlength{\parindent}{0pt} % <-- removes the indent
% \renewcommand\thefootnote{*}
% \footnotetext{These authors contributed equally as first authors and reserve the right to list their names first in their respective versions.}
% \renewcommand\thefootnote{†}
% \footnotetext{Corresponding author.}
% \endgroup

\begin{abstract}
Generating editable 3D CAD models from natural language remains challenging, as existing text-to-CAD systems either produce meshes or rely on scarce design-history data. We present NURBGen, the first framework to generate high-fidelity 3D CAD models directly from text using Non-Uniform Rational B-Splines (NURBS). To achieve this, we fine-tune a large language model (LLM) to translate free-form texts into JSON representations containing NURBS surface parameters (\textit{i.e}, control points, knot vectors, degrees, and rational weights) which can be directly converted into BRep format using Python. We further propose a hybrid representation that combines untrimmed NURBS with analytic primitives to handle trimmed surfaces and degenerate regions more robustly, while reducing token complexity. Additionally, we introduce partABC, a curated subset of the ABC dataset consisting of individual CAD components, annotated with detailed captions using an automated annotation pipeline. NURBGen demonstrates strong performance on diverse prompts, surpassing prior methods in geometric fidelity and dimensional accuracy, as confirmed by expert evaluations. Code and dataset will be released publicly.

% Generating editable 3-D CAD geometry from natural language remains an open challenge: existing text-to-3-D systems either output surface meshes or require scarce design-history supervision, making them unsuitable for real-world engineering workflows. We introduce NURBGen, the first framework that translates free-form text into native, parametric CAD models by predicting complete Non-Uniform Rational B-Spline (NURBS) surface specifications—control points, knot vectors, degrees, and rational weights—that are immediately convertible to boundary-representation solids. Because NURBGen is history-agnostic, it can exploit large shape corpora that lack feature trees. To that end, we created partABC, a subset of the ABC dataset augmented with automatically generated, fine-grained captions, yielding the largest training resource to date for text-driven CAD synthesis. Trained on partABC, NURBGen produces geometrically precise and semantically faithful models across diverse prompts and significantly outperforms mesh-based and design-history baselines in expert preference studies. Both the NURBGen codebase and the partABC dataset will be released to catalyze future research on language-conditioned, parametric 3-D design.

    % To bridge the gap between discrete token prediction and continuous geometry, we introduce a hybrid training objective combining cross-entropy loss with a numerical consistency loss, ensuring both syntactic correctness and geometric fidelity.
\end{abstract}
\section{Introduction}

\noindent Computer-Aided Design (CAD) plays a fundamental role in modern engineering, product design, and digital manufacturing workflows~\cite{cad_additive_review, cad_status_engineering}. It enables precise, parametric modeling of complex mechanical and architectural components. However, creating detailed CAD models typically requires expert knowledge of professional design software-such as Onshape \url{(https://www.onshape.com)} or AutoCAD \url{(https://www.autodesk.com/products/autocad/overview)}, and remains a labor-intensive and, time-consuming task.

% \footnote{Popular CAD tools include Onshape (\url{https://www.onshape.com}), AutoCAD (\url{https://www.autodesk.com/products/autocad/overview}), and SolidWorks (\url{https://www.solidworks.com})}
\vspace{1mm}

\noindent Researchers have therefore proposed deep learning-based approaches for automatic CAD modeling from high-level inputs such as natural language~\cite{text2cad}, images~\cite{img2cad}, or point clouds~\cite{point2cad}. Among these, text-to-CAD generation offers a simple, intuitive interface that allows designers to describe 3D objects in natural language, bypassing the need for expert modeling skills. However, nearly all prior methods~\cite{cadllama, cadgpt, text2cad} rely on design-history-based representations~\cite{deepcad, cadsignet, cadparser}, where shapes are constructed via sequences of parametric operations—extrusions, and 2D sketches. While intuitive and highly editable, these methods are trained on small-scale datasets like DeepCAD~\cite{deepcad}, which mostly contain low-complexity parts (e.g., cuboids, cylinders), limiting generalization in real-world scenarios.
\vspace{1mm}

\begin{figure}[t]
    \centering
    \includegraphics[width=0.96\linewidth]{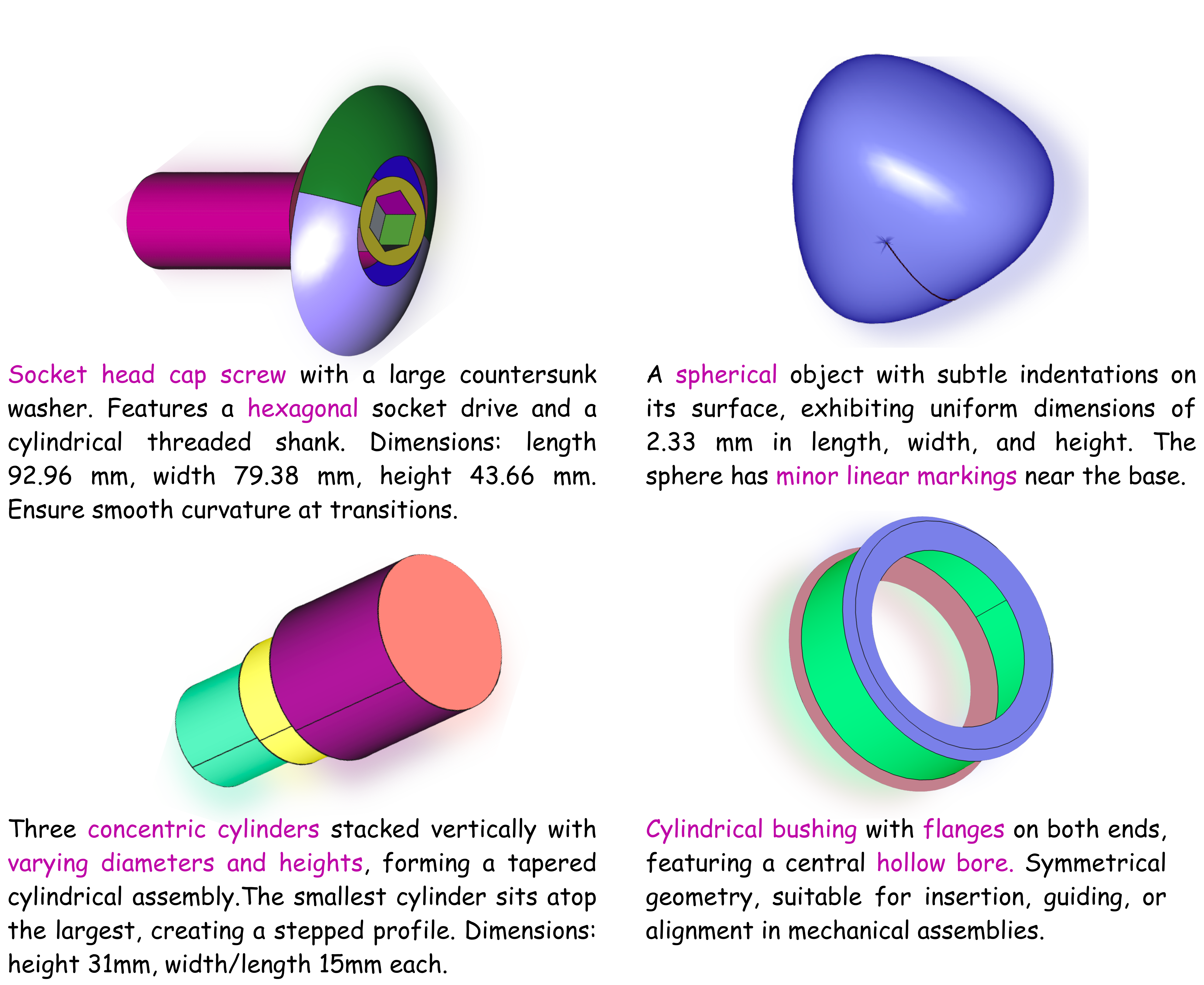}
    \caption{Text-to-CAD generation results from NURBGen, showcasing reconstructed CAD models from text prompts.}
    \label{fig:teaser}  
    % \vspace{-1mm}
\end{figure}

\noindent In contrast, the ABC dataset~\cite{abc}, which contains over a million 3D CAD models, remains comparatively underutilized in text-to-CAD research due to two key limitations. First, ABC represents geometry in Boundary Representation (BRep) form, which lacks design history. BReps define solids using analytic surface patches—most commonly NURBS, the industry standard for their precision and parametric control. However, NURBS-based modeling is rarely explored in deep generative research due to the challenge of efficient representation~\cite{nurbot}, non-differentiability of knot vectors~\cite{nurbsdiff}, high parameter variability, and trimming complexity inherent to NURBS geometry. NeuroNURBS~\cite{neuronurbs} partially addresses this by learning latent codes for untrimmed NURBS surfaces via a non-autoregressive transformer VAE, but it does not support language-based generation and cannot model complex shapes due to the trimming issues. Second, ABC lacks high-quality text descriptions, making it difficult to train or evaluate text-conditioned generative models.
\vspace{1mm}

\noindent In this work, we present \textbf{NURBGen}, the first framework for generating 3D CAD models from natural language using structured, symbolic NURBS representations. Unlike prior work that learns dense latent codes~\cite{neuronurbs}, we treat each NURBS surface as a language-aligned object: a sequence of tokens encoding control points, degrees, weights, and knot vectors in JSON format. This allows us to formulate text-to-CAD as a language modeling task. We fine-tune a large language model (Qwen3-4B) to map textual descriptions to these NURBS parameters, producing outputs that are editable and directly compatible to BRep format. To support this, we construct \textbf{partABC}, a curated dataset of more than 300k part-level CAD models from the ABC dataset, each represented as a sequence of NURBS surfaces and serialized with manageable context lengths ($\leq$ 8k tokens). We also generate high-quality natural language descriptions of the CAD models using an automatic annotation pipeline for the supervised fine-tuning task.
\vspace{1mm}

\noindent A key design choice to manage context length is our use of untrimmed NURBS surfaces similar to NeuroNURBS~\cite{neuronurbs}. However, this introduces a limitation it cannot capture trimmed geometry precisely. To address this, we propose a hybrid symbolic representation that replaces NURBS with primitive analytic curves (e.g., circles, B-splines, arcs, and lines) to accurately model such faces. This maintains the structural format required for LLM fine-tuning and inference. Our experiments demonstrate that NURBGen can outperform the current state-of-the-art methods in high-fidelity text-to-CAD generation as shown in Figure~\ref{fig:teaser}. Our contributions can be summarized as follows
\begin{itemize}
    \item We propose NURBGen, the first framework for LLM-driven NURBS-based text-to-CAD framework.
    \item We introduce partABC, a large-scale multi-modal dataset of 300k CAD parts from the ABC dataset with NURBS annotations and high-quality captions using an automatic annotation pipeline.
    \item We design a hybrid representation combining untrimmed NURBS with analytic primitives to accurately model trimmed and degenerate surfaces while maintaining structural compatibility for LLM fine-tuning.
    \item Our extensive experiments demonstrate NURBGen's superior performance over existing baselines.
\end{itemize}

\section{Related Work}

\noindent \textbf{CAD Generation:} Earlier approaches to CAD generation primarily focused on low-level geometry tasks such as surface fitting~\cite{parsenet, point2cad}, point cloud classification~\cite{pointnet, pointnet++}, BRep segmentation~\cite{cadopsnet, brepgat, sharp2023}, or BRep structure prediction~\cite{complexgen, brepdetnet}, rather than generating fully parametric CAD models. A major shift came with DeepCAD~\cite{deepcad}, which introduced a design-history-based representation where CAD models are expressed as sequences of 2D sketches and 3D operations (e.g., extrusions). This formulation enabled sequence-to-sequence modeling of CAD generation. Building on this, later works explored cross-modal CAD synthesis from point clouds~\cite{cadsignet, transcad, cadrecode}, images~\cite{img2cad}, natural language~\cite{cadllama, cad-instruct, text2cad, cadtranslator, cadmium, text-to-cad}, or combinations thereof~\cite{cadrille, cad-mllm}. While design-history representations are highly interpretable and editable, their reliance on proprietary CAD operation data presents a major bottleneck for large-scale public research. Public datasets like DeepCAD-170k~\cite{deepcad}, Fusion360-8k~\cite{fusion360}, and CADParser-50k~\cite{cadparser} are limited in size and complexity, often consisting of simple, synthetic parts~\cite{cadmium}, which restricts generalization to real-world scenarios. Alternative approaches for CAD generation operate directly on BRep geometry~\cite{brepnet} or leverage SDF supervision~\cite{extrudenet, secadnet, caprinet}. However, these methods don't generalizes well. Recent work BrepGen~\cite{brepgen}, for example, generates BRep topology including vertices, edges, and faces via a hierarchical latent diffusion model. In contrast, we represent BReps as sequences of structured NURBS surfaces, allowing us to frame text-to-CAD as a language generation task. This enables fine-tuning an LLM on partABC, which is more diverse and larger than those used in prior work.
\vspace{1mm}
\begin{figure*}[ht]
    \centering
    \includegraphics[width=1\linewidth]{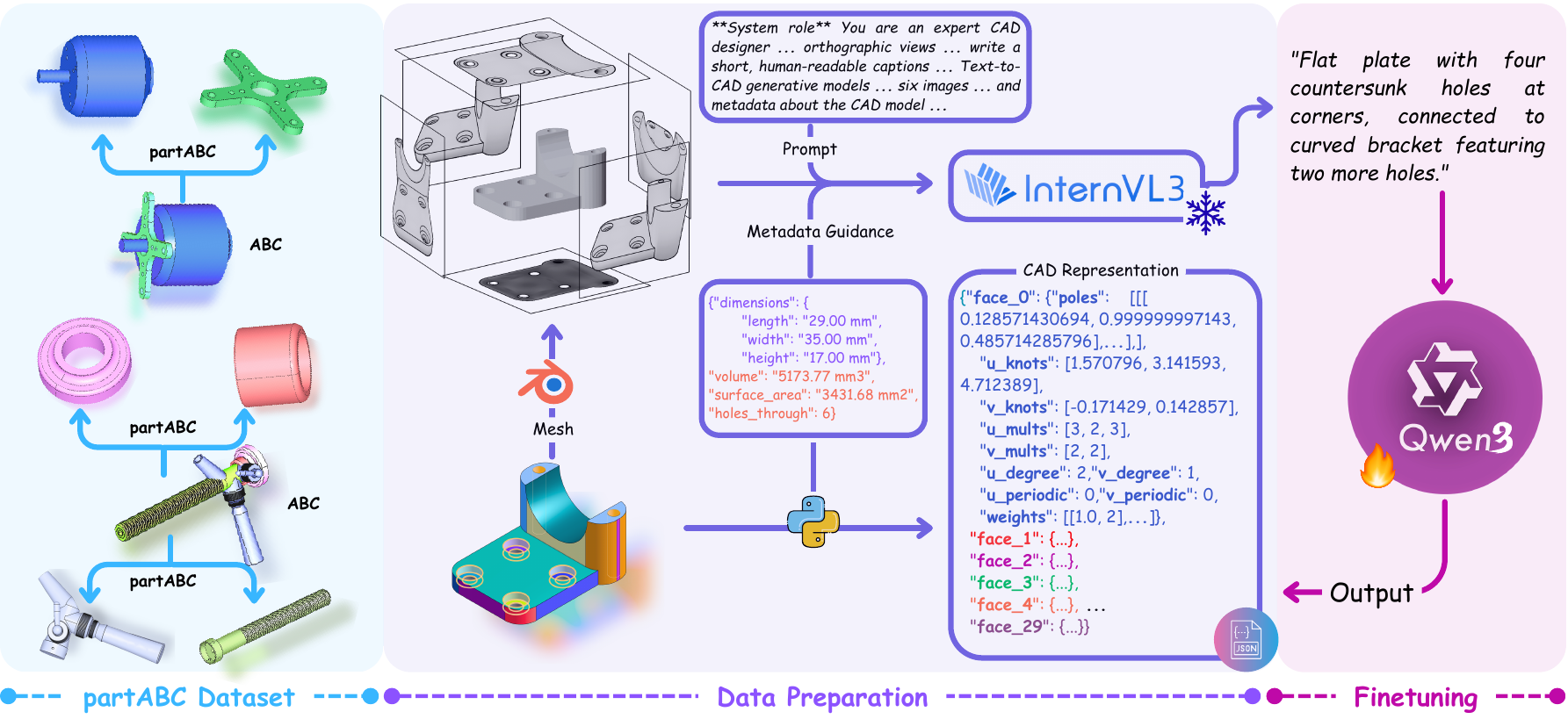}
    \caption{Overview of our partABC dataset, data preparation and fine-tuning pipeline. \textbf{Left:} We extract part-level CAD models from the ABC dataset by decomposing CAD assemblies into individual components. \textbf{Middle:} Each part is represented using a hybrid format—faces are encoded as untrimmed NURBS surfaces, with analytic primitives used where NURBS fitting fails. We also generate high-quality captions using InternVL3-13B with a metadata-guided annotation pipeline. \textbf{Right:} We fine-tune Qwen3-4B to map text captions to structured hybrid CAD representations, which can be directly converted to BRep models.}
    \label{fig:data-prep-pipeline}
\end{figure*}  

\noindent \textbf{Nurbs Modeling:} The adoption of analytic surfaces like NURBS in learning-based systems remained limited~\cite{nurb_survey,nurb_optimization}. NURBSDiff~\cite{nurbsdiff} introduced differentiable NURBS fitting for geometry optimization and reconstruction. \cite{drpg} proposed differentiable rendering of NURBS surfaces for inverse graphics tasks. The most relevant prior work is NeuroNURBS~\cite{neuronurbs}, which encodes untrimmed NURBS surfaces using a non-autoregressive transformer autoencoder into latent vectors for supporting tasks like reconstruction or segmentation, but not text-conditioned generation. However, the exclusive use of untrimmed NURBS surfaces limits generalization, as not all CAD models can be accurately represented without trimming. To address this, we adopt a hybrid strategy: while untrimmed NURBS serve as our primary representation, we replace them with analytic primitives such as lines, arcs, and B-splines for faces where NURBS fitting fails.
\vspace{1mm}

\noindent \textbf{LLM for 3D Generation}: LLMs have been widely adopted across domains such as robotics~\cite{llm_robotics} and 3D scene understanding or grounding~\cite{pointllm, 3dllm}. Their application to 3D generation is a relatively new but promising direction, as LLMs offer strong spatial priors and multimodal reasoning capabilities. A central challenge, however, lies in encoding 3D geometry into a sequential format compatible with language modeling. Recent work such as LLaMA-Mesh~\cite{llama_mesh} finetunes LLaMA~\cite{llama3} to generate mesh vertices and faces as plain text, demonstrating the potential of autoregressive text-based 3D synthesis. In the CAD domain, existing LLM-driven text-to-CAD methods primarily rely on design-history-based representations~\cite{cadllama, cadrecode, cad-mllm, llm_cad_survey}. However, these methods are constrained by the scarcity and simplicity of the public datasets. In contrast, our method introduces a structured NURBS-based representation that enables symbolic, surface-level generation. This formulation aligns naturally with language modeling and allows us to leverage the large-scale and geometrically diverse partABC dataset for fine-tuning LLMs for the text-to-CAD task.

\section{Background}\label{sec:background}

Before presenting our method, we briefly review the fundamentals of NURBS.
Non-Uniform Rational B-Splines (NURBS) are the standard representation for curves and surfaces in CAD and geometric modeling. They extend B-splines by assigning weights to control points, enabling both free-form and analytic shapes (e.g., circles, ellipses). Their compactness, smoothness, and precise parametric control make them central to modern CAD systems. A NURBS curve of degree $p$ is defined by:
\begin{itemize}
    \item A set of $n+1$ control points $\{ \mathbf{P}_i \in \mathbb{R}^d \}_{i=0}^{n}$,
    \item A corresponding set of weights $\{ w_i \in \mathbb{R}^+ \}_{i=0}^{n}$,
    \item Knot vector $\mathbf{U} = \{ u_0, \ldots, u_{n+p+1} \}$, $u_i\geq u_j, \forall i>j $
    \item Basis functions $N_{i,p}(u)$ defined recursively.
\end{itemize}
\noindent The NURBS curve is then given by:
\begin{align}
    \mathbf{C}(u) = \frac{\sum_{i=0}^{n} N_{i,p}(u) w_i \mathbf{P}_i}{\sum_{i=0}^{n} N_{i,p}(u) w_i}, \quad u \in [u_p, u_{n+1}]
\end{align}

The B-spline basis functions $N_{i,p}(u)$ are defined recursively using the Cox-de Boor formula:
\begin{equation}\label{eq:basis}
\begin{aligned}
N_{i,0}(u) &=
\begin{cases}
1 & \text{if } u_i \leq u < u_{i+1}, \\
0 & \text{otherwise},
\end{cases} \\
N_{i,p}(u) &= 
\frac{u - u_i}{u_{i+p} - u_i} \, N_{i,p-1}(u) \\
&\quad + \frac{u_{i+p+1} - u}{u_{i+p+1} - u_{i+1}} \, N_{i+1,p-1}(u)
\end{aligned}
\end{equation}

\noindent A NURBS surface is defined similarly, as the tensor product of two NURBS curves in parameters $u$ and $v$. Given control points $\mathbf{P}_{ij}$, weights $w_{ij}$, knot vectors $\mathbf{U}$ and $\mathbf{V}$, and degrees $p$, $q$, the NURBS surface is:
\begin{align}
\mathbf{S}(u, v) &= 
\frac{
    \sum_{i=0}^{n} \sum_{j=0}^{m} N_{i,p}(u) M_{j,q}(v) w_{ij} \mathbf{P}_{ij}
}{
    \sum_{i=0}^{n} \sum_{j=0}^{m} N_{i,p}(u) M_{j,q}(v) w_{ij}
},  \\
&\quad (u, v) \in [u_p, u_{n+1}] \times [v_q, v_{m+1}] \notag
\end{align}

\noindent Here, $N_{i,p}(u)$ and $M_{j,q}(v)$ are the B-spline basis functions as defined in Eq.~\ref{eq:basis} in the $u$- and $v$-directions, respectively.

\section{Data Preparation}\label{sec:data_prep}
In this section, we describe our data preparation pipeline, illustrated in Figure~\ref{fig:data-prep-pipeline} (left and middle column). Our objective is to extract a NURBS-based surface representation for a BRep model in JSON format, along with a high-quality textual caption, which will serve as supervision to fine-tune an LLM for precise and editable text-to-CAD generation. To this end, we construct a new dataset, \textbf{partABC}, derived from the unlabeled, assembly-level ABC dataset. The following subsections detail our NURBS representation format and explain the motivation and processing steps used to build partABC.

\subsection{1. CAD Representation}
A BRep solid models geometry as a collection of topologically connected faces, each defined by a bounded parametric surface. In modern CAD systems, these surfaces are most commonly represented using NURBS surfaces due to their ability to accurately model both analytic primitives (e.g., planes, cylinders, tori) and complex free-form geometry with high continuity and compactness. To reconstruct a BRep solid in a symbolic and editable form, it is essential to extract the full set of NURBS surface parameters for each face. Using pythonOCC~\footnote{https://github.com/tpaviot/pythonocc-core}, we propose a robust pipeline for converting BReps into parametric NURBS representation. 

% We use pythonOCC~\footnote{https://github.com/tpaviot/pythonocc-core}library for our data processing task.

\vspace{1mm}
\noindent Given a BRep solid, we begin by normalizing the geometry to fit within a $2\times2\times2$ bounding box centered at the origin, ensuring consistent scale and alignment across all samples. We then apply \texttt{BRepBuilderAPI\_NurbsConvert} to convert each face into its untrimmed NURBS representation. This step standardizes all underlying analytic and freeform surfaces—such as planes, cylinders, and spline patches—into rational B-splines, providing a uniform surface representation. Next, we traverse each face using \texttt{TopExp\_Explorer} and extract its surface parameters via the \texttt{Geom\_BSplineSurface} API. For each face, we retrieve the control points (also called poles), knot vectors in both parametric directions, knot multiplicities, degrees in u and v, rational weights, and periodicity flags. Knot multiplicities specify how many times each knot value appears in the knot vector. Periodicity flags indicate whether the surface is seamlessly closed in the u and/or v direction, as in cylindrical or toroidal geometries. With all these parameters extracted, the original surface can be exactly reconstructed using the \texttt{Geom\_BSplineSurface} constructor.
\vspace{1mm}
% \begin{figure}[t]
%     \centering
%     \includegraphics[width=1\linewidth]{AnonymousSubmission/LaTeX/main/assets/hybrid_representation.pdf}
%     \caption{Our proposed hybrid representation. \textbf{Left:} Untrimmed NURBS surfaces introduce artifacts in hole-like or thin regions. \textbf{Right} We resolve this by substituting their NURB representation with analytic curves (\textit{e.g.}, lines, circles) for improved geometric fidelity.}
%     \label{fig:hybrid}
% \end{figure}
\begin{figure}[t]
    \centering
    \includegraphics[width=1\linewidth]{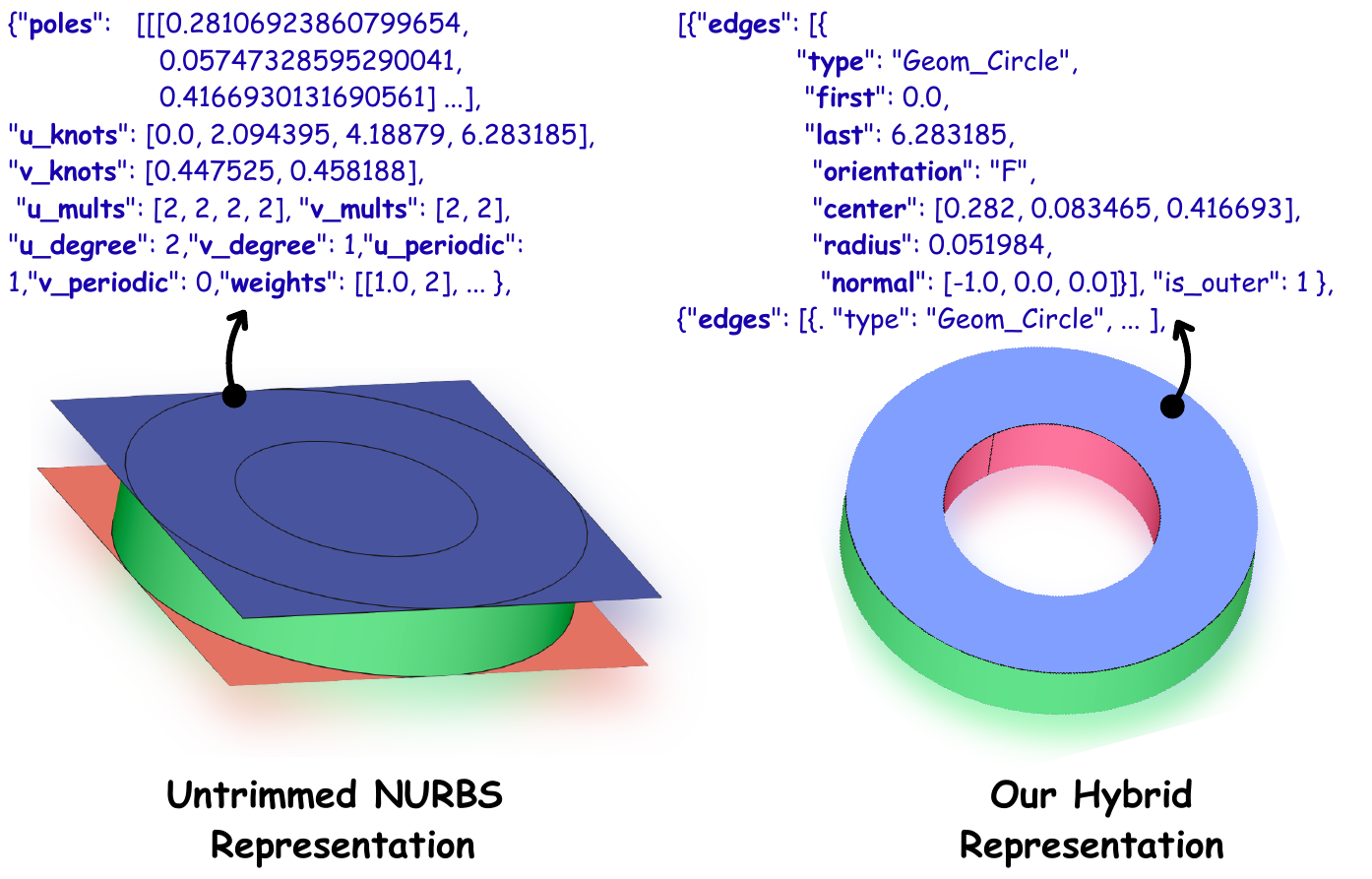}
    % \vspace*{-0.6\baselineskip}
    \caption{Our proposed hybrid representation. \textbf{Left:} Untrimmed NURBS surfaces introduce artifacts in hole-like or thin regions. \textbf{Right} We resolve this by substituting their NURB representation with analytic curves (\textit{e.g.}, lines, circles) for improved geometric fidelity.}
    % \vspace*{-0.6\baselineskip}
    \label{fig:hybrid}
\end{figure}
\begin{figure*}[ht]
    \centering
    \includegraphics[width=1\linewidth]{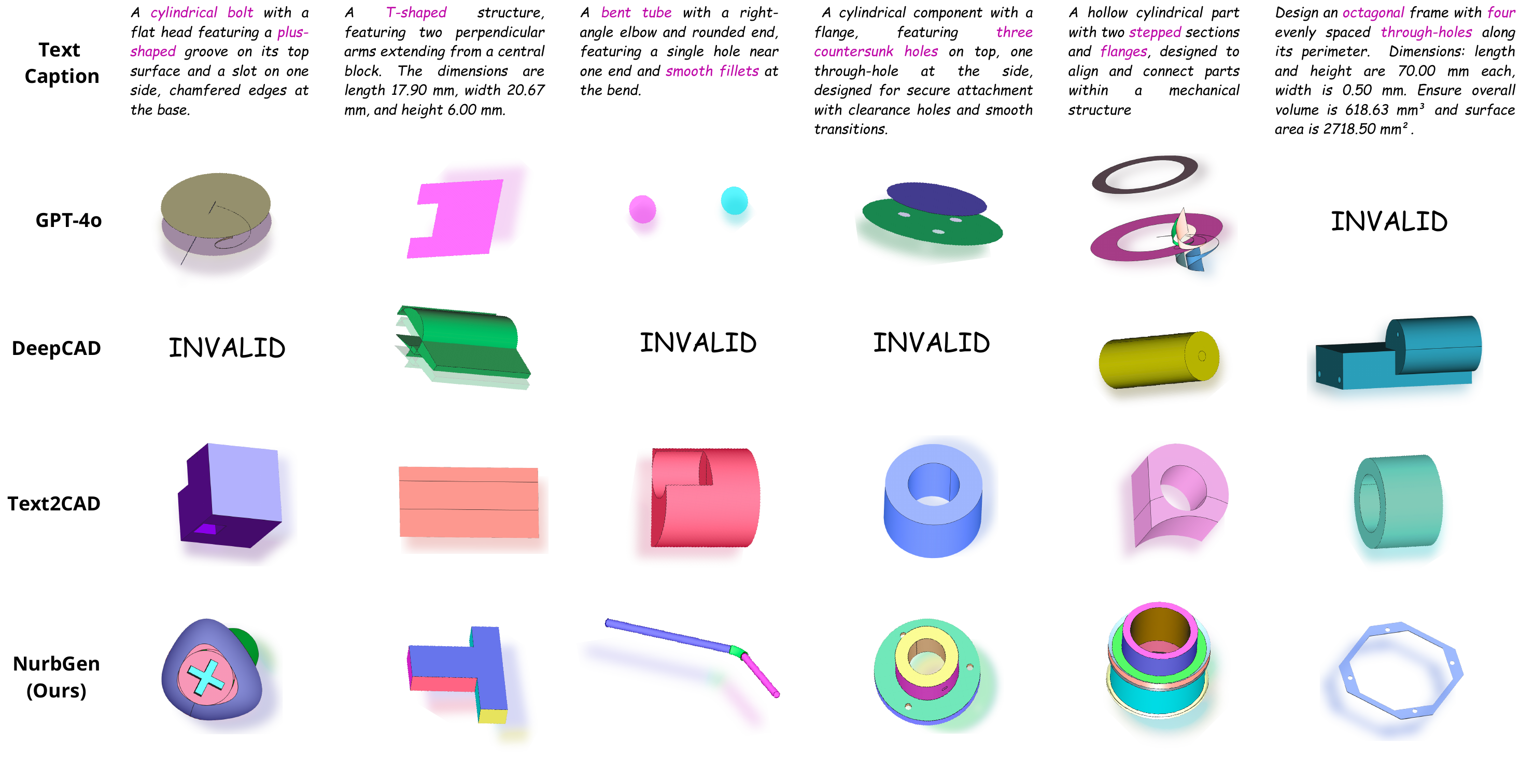}
    % \vspace*{-0.6\baselineskip}
        \caption{Qualitative comparison of reconstructed CAD models from text prompts. From top to bottom, we show generations from GPT-4o, DeepCAD, Text2CAD, and our proposed NURBGen. NURBGen consistently produces more detailed and structurally coherent results, with higher fidelity to the input prompt and fewer geometric artifacts compared to baselines.}
    \label{fig:qualitative}
\end{figure*}

\noindent However, Not all surfaces can be robustly represented by untrimmed NURBS. In particular, thin regions around holes or fillets often introduce geometric artifacts or reconstruction errors (see Fig.~\ref{fig:hybrid}). To address such degenerate cases, we adopt a hybrid representation: instead of enforcing a NURBS fit, we revert to simpler analytic primitives such as lines, circles, B-splines, ellipses, parabolas, and hyperbolas. These primitives are extracted from the original BRep faces prior to NURBS conversion. We detect degenerate or poorly reconstructed faces by comparing each reconstructed surface $f_n$ with its ground-truth counterpart $f_{gt}$ using the Chamfer Distance (CD) between their sampled point clouds:
\begin{equation}
CD(f_n, f_{gt}) \leq \epsilon
\end{equation}
\noindent $CD$ measures the average squared distance from points in one set to their nearest neighbors in another. If it is below a threshold $\epsilon$, the NURBS approximation is deemed acceptable; otherwise, we retain the original analytic primitive. We empirically set $\epsilon=6\times 10^{-4}$.
\vspace{1mm}

\noindent We represent each face using either its extracted NURBS parameters or its analytic primitive definition, based on reconstruction quality. In practice, about 70\% of faces are modeled using NURBS, while 30\% fall back to analytic primitives. This hybrid representation stored in structured JSON offers a more expressive and compact alternative to the purely NURBS-based format used in NeuroNURBS~\cite{neuronurbs}. While analytic primitives can represent simple geometry, they lack the flexibility to capture free-form surfaces. NURBS, on the other hand, provide a unified framework that can model both standard analytic shapes and complex free-form surfaces within a single patch—often replacing multiple primitives such as segmented arcs or partial cylinders. By combining both representations, our hybrid approach improves robustness and reduces parameter count for simpler shapes, resulting in shorter and more token-efficient inputs for LLM fine-tuning.

% \vspace*{-0.3\baselineskip}
\subsection{2. Annotation Pipeline}
Supervised fine-tuning of our text-to-CAD model requires paired textual descriptions, but the ABC dataset lacks captions. To address this, we design an automated annotation pipeline using a VLM to generate high-quality captions for CAD models at scale as shown in Figure.~\ref{fig:data-prep-pipeline} (middle).
\vspace{1mm}

\noindent \textbf{Multi-View Rendering:} Each BRep is first converted into a textureless triangular mesh and rendered from six viewpoints at a resolution of $512 \times 512$ using Blender~\footnote{\url{https://docs.blender.org/api/3.6/index.html}}. Four of the camera views follow the orientation strategy proposed in~\citepalias{marvel}, while the remaining two capture the top and bottom perspectives. To enhance geometric perception and visual clarity, we enable Blender’s Freestyle renderer to overlay clean silhouette and edge contours on each image.
\vspace{1mm}

% \noindent \textbf{Metadata Guidance:} High-quality captions for CAD models should go beyond simple naming and include critical geometric features like the count of holes or global properties (e.g., overall dimensions). Text2CAD~\cite{text2cad} uses minimal JSON-based design history to guide VLM, while recent work like MARVEL~\citepalias{marvel} uses metadata to enable fine-grained 3D annotations. Inspired by these, we extract information that is typically difficult for VLMs to infer—such as length, width, height, surface area, volume, and the number of through-holes. This metadata is then injected into the annotation prompt used in our caption generation process.
% \vspace{1mm}

\noindent \textbf{Metadata Guidance for Caption Generation:} High-quality captions for CAD models should go beyond simple naming and incorporate essential geometric features such as the number of through-holes, overall dimensions, surface area, and volume. Previous work like Text2CAD~\cite{text2cad} leverages minimal JSON-based design history to guide vision-language models (VLMs), while MARVEL~\citepalias{marvel} uses structured metadata for fine-grained 3D annotation. Building on these ideas, we extract geometric metadata that is often inaccessible to VLMs—specifically, length, width, height, surface area, volume, and the number of topological holes (genus). We compute overall dimensions by fitting an axis-aligned bounding box to the CAD geometry using OpenCascade’s \texttt{Bnd\_Box}. Volume and surface area are obtained using OpenCascade’s built-in mass property computation (\texttt{brepgprop.VolumeProperties} and \texttt{SurfaceProperties}). 

\vspace{0.1cm}
\noindent To estimate the number of through-holes, we first generate a watertight mesh from BRep. We then compute the Euler characteristic, $\chi = V - E + F$, where $V, E, F$ are the mesh vertices, edges, and faces. Using the Euler–Poincaré formula for closed 2-manifolds, the genus is given by $g = 0.5 \times (2 - \chi)$, which corresponds to the number of topological through-holes~\cite{cibulkajoint}. This metadata is then injected into the annotation prompt, which guides the VLM to generate captions with precise measurements.

\vspace{1mm}

\begin{figure}[t]
    \centering
    \includegraphics[width=0.96\linewidth]{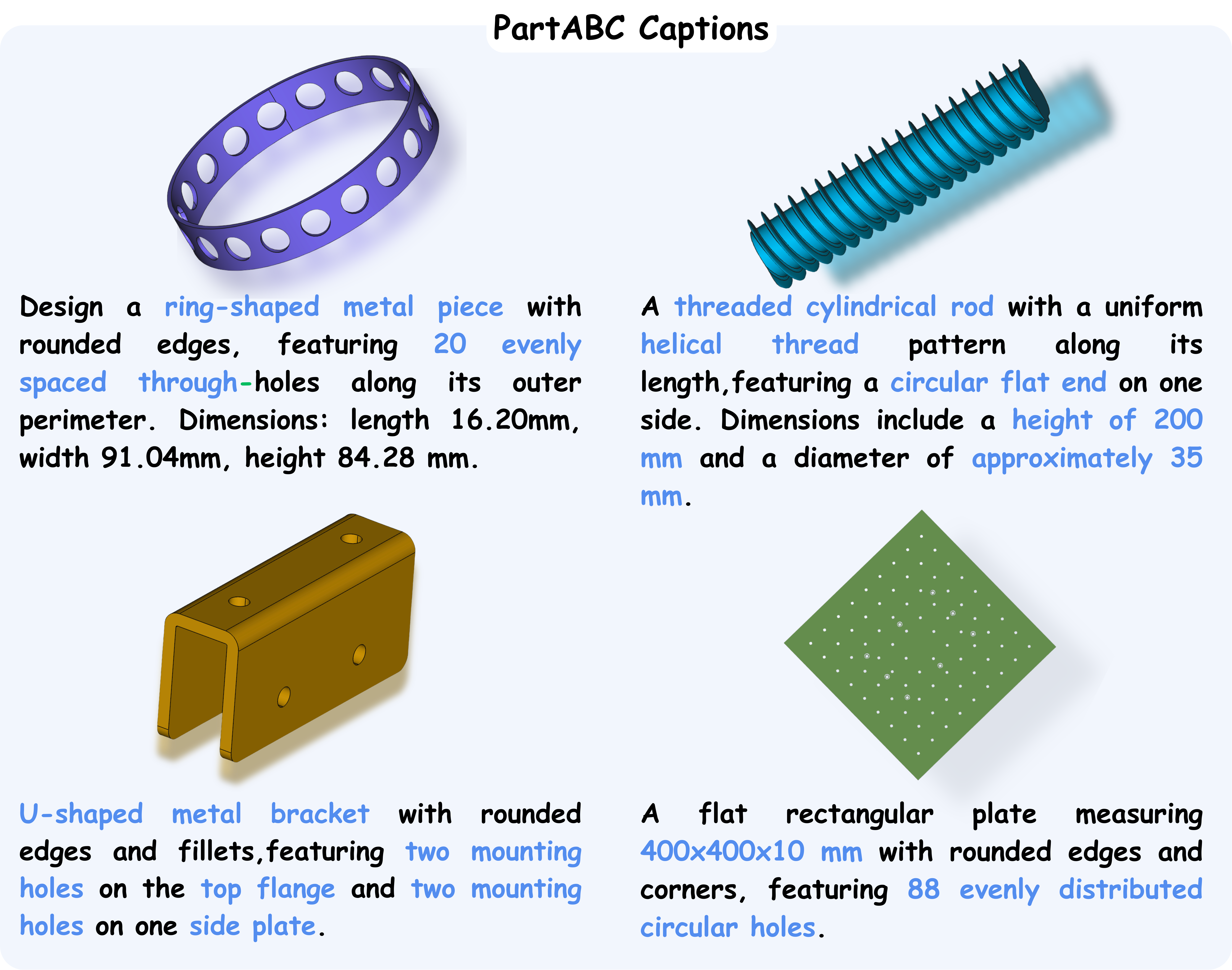}
    \caption{Captions from partABC dataset generated using our captioning pipeline.}
    \label{fig:caption}
\end{figure}

\noindent \textbf{Caption Generation:} We use InternVL3-13B~\cite{internvl3}, a multi-view VLM, which takes six rendered views of the CAD model along with the metadata-augmented annotation prompt as input. It then processes multi-view images of the CAD model simultaneously to generate a coherent and geometry-aware caption. Rather than focusing solely on object category names, we prioritize shape-centric descriptions that capture structural characteristics (\textit{"a bent tube.."},\textit{a flat cylindrical bolt with six holes..}). The inclusion of dimensional metadata and hole counts further grounds the captions in precise geometric details, resulting in more informative and reliable annotations as shown in Figure~\ref{fig:caption}.

\vspace*{-0.5\baselineskip}
\subsection{3. partABC Dataset}
In this section, we describe the construction of the partABC dataset as shown in Figure~\ref{fig:data-prep-pipeline} (Left Column). While our data processing pipeline supports any CAD model, we focus on the ABC dataset due to its large scale and geometric diversity. ABC contains 1M CAD models, but processing the full set is computationally expensive and time-consuming. Therefore, we limit our preprocessing to 200k models for this project. However, many of these are assembly-level designs with a large number of faces, resulting in JSON representations that can exceed 100k tokens well beyond the context window and training budget of our project. To address this, we leverage the fact that BReps in ABC often encode part-level substructures within these assemblies. Using PythonOCC, we programmatically extract these individual parts, each representing a self-contained and geometrically coherent component. From the 200k processed assemblies, this provides us with 3M part-level CAD instances.
\vspace{1mm}

\begin{figure}[t]
    \centering
    \includegraphics[width=0.95\linewidth]{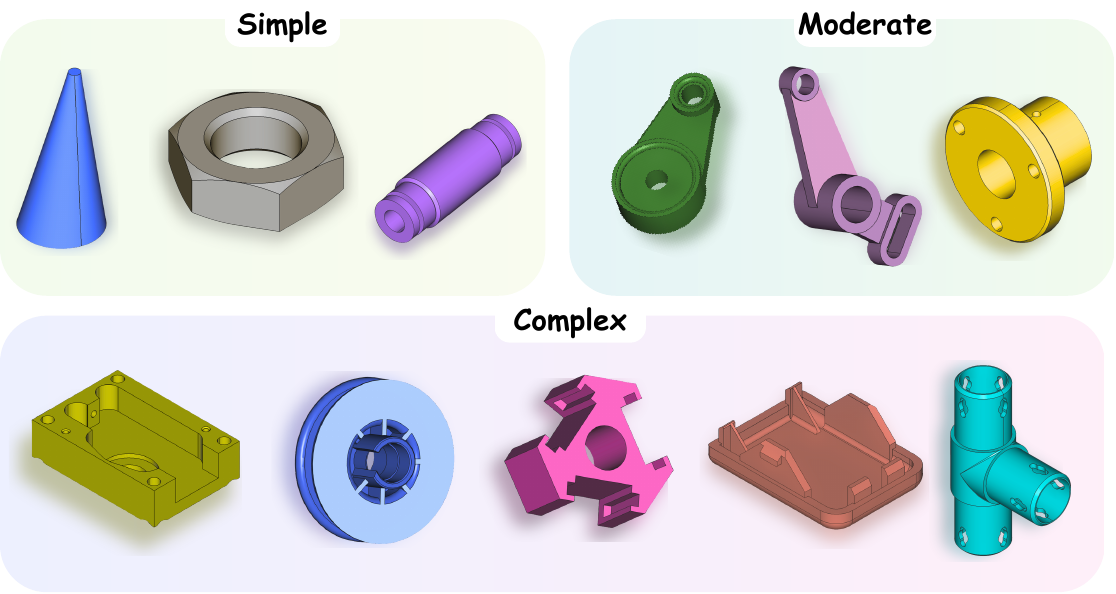}
    \caption{Example CAD parts from the partABC dataset across complexity tiers- simple (top-left), moderate (top-right), and complex (bottom).}
    \label{fig:complexity}
\end{figure}

\noindent \textbf{Filtering:} However, extracting part-level CAD models from larger assemblies introduces a key challenge: many of the resulting shapes tend to be geometrically simple such as cuboids or cylinders. This can lead to an imbalanced training set and bias the fine-tuned LLM toward generating trivial geometry. To address this, we apply a complexity-aware filtering strategy using a weighted scoring function that prioritizes geometrically rich and structurally diverse parts. Each part-level model is scored using
% \vspace*{-0.5\baselineskip}
\begin{align} \notag
\texttt{w}(B) = {} & 
    l_1 \times \texttt{token\_count} +  l_2 \times \texttt{through\_holes} \notag \\
    & + l_3 \times \frac{\texttt{surface\_area}}{\texttt{volume}} + l_4 \times \texttt{bbox\_diag} \notag
\end{align}
where $l_1=0.35, l_2=0.3, l_3=0.25, l_4=0.1$ are selected using empirical experiment on 100 samples. \texttt{token\_count} refers to the size of tokens after tokenizing the JSON using Qwen3 Tokenizer~\cite{qwen3}, \texttt{through\_holes} counts the number of holes that pass through the entire part, and \texttt{bbox\_diag} is the length of the diagonal of the part’s axis-aligned bounding box. Based on $\texttt{w}(B)$, we categorize parts into simple ($\leq 0.12$), moderate ($0.12$–$0.23$), and complex ($>0.23$) tiers.
From 3M extracted parts, we retain 10\% simple, 50\% moderate, and 40\% complex models, forming the final partABC dataset of $\sim$300k high-quality samples as shown in Figure~\ref{fig:complexity}.

\section{Experimental Results}
% \vspace{0.3mm}
In this section, we provide details of our experiments and discuss evaluation results with baselines.
\vspace{1mm}

\noindent \textbf{Datasets:} We use the curated partABC dataset for supervised fine-tuning of our model, with 95\%–2.5\%-2.5\% split for training, validation, and testing. To reduce context length and improve token efficiency, we round the NURBS control point coordinates to $6$ decimal places. Additionally, we compress control point weights using a \textit{(value, frequency)} representation scheme.

\vspace{1mm}

\noindent \textbf{Implementation Details:} We fine-tune Qwen3-4B model~\cite{qwen3} using AdamW~\cite{adamw} with a learning rate of $5{\times}10^{-5}$ and linear warm-up. LoRA~\cite{lora} is applied with rank $64$ and $\alpha{=}128$. Training runs for $180$k steps with batch size $1$ on $4{\times}$H200 GPUs over 3 days. The context window is $8192$ during training and $14$k during inference, with temperature $0.3$. On RTX 3090, the model achieves a generation throughput of $\sim 800$ tokens per second. Figure~\ref{fig:data-prep-pipeline} (Right column) shows the finetuning task.
\vspace{1mm}

\noindent \textbf{Baselines:} We compare against strong open-source baselines for text-to-CAD generation. While recent models like CAD-LLaMA~\cite{cadllama} and CADFusion~\cite{text-to-cad} report promising results, their implementations are not publicly available. Moreover, to the best of our knowledge, there are no open-source models capable of generating NURBS-based CAD representations from text. Hence, we focus our comparison with open-source methods, including Text2CAD~\cite{text2cad}, DeepCAD~\cite{deepcad}, and GPT-4o. We use official pretrained weights for Text2CAD and retrain DeepCAD for 100 epochs following the Text2CAD protocol. GPT-4o is evaluated using 2-shot prompting with example caption–JSON pairs. 
% All methods are tested on 100 randomly selected prompts from our test dataset.

\vspace{1mm}

% \noindent \textbf{Metrics:} Similar to the text-to-3D domain~\cite{benchmark_3d}, the text-to-CAD task lacks standardized benchmarks, as multiple valid CAD outputs can correspond to the same prompt~\cite{text2cad}. To evaluate prompt fidelity and reconstruction quality, we conduct both human and GPT-4o assessments. For the human study, we recruit five CAD designers with varying expertise levels and ask them to choose the model that best matches each input prompt. We report top-1 voting results based on majority preference. For GPT-4o evaluation, we provide a $2\times2$ grid of multi-view images and the prompt, asking it to select the most faithful reconstruction. If the compared outputs are visually and semantically similar, both reviewers return ‘Undecided.’ We also report the Invalidity Ratio, defined as the proportion of generated outputs that fail to reconstruct into valid BRep models.

% Similar to the text-to-3D domain~\cite{benchmark_3d}, the text-to-CAD task lacks standardized benchmarks, as multiple valid CAD outputs can correspond to the same prompt~\cite{text2cad}

\noindent \textbf{Metrics:} We evaluate both geometric fidelity and visual alignment of generated CAD models.
For geometry, we compute Chamfer Distance (CD), Hausdorff Distance (HD), Jensen–Shannon Divergence (JSD), and Minimum Matching Distance (MMD) on 7,500 test samples, using 8,192 uniformly sampled points normalized within a unit cube. For visual evaluation, we measure prompt fidelity using both human and GPT-4o preference studies on 1k and 5k samples, respectively. In the human study, five CAD designers of varying expertise select the reconstruction that best matches each input prompt, with majority voting reported as Top-1 accuracy. For GPT-4o evaluation, we present a $2\times2$ grid of multi-view renderings along with the prompt, asking it to choose the most faithful reconstruction or mark as “Undecided” if outputs are visually comparable. Finally, we report the Invalidity Ratio (IR), the percentage of generated models that fail to convert into valid B-Rep structures.

% In addition, we evaluate geometric and distributional metrics on the full 7,500-sample test set, including Chamfer Distance (CD), Hausdorff Distance (HD), Jensen–Shannon Divergence (JSD), and Minimum Matching Distance (MMD), all scaled by 100. These quantitative metrics complement the human and GPT-based evaluations and further substantiate the comparative performance of all methods.

% The final score is averaged. Additionally, since some of our prompts include explicit dimension information, we compute the mean squared error (MSE) on those samples to quantitatively evaluate dimensional accuracy.
\begin{figure}[t]
    \centering
    \includegraphics[width=0.95\linewidth]{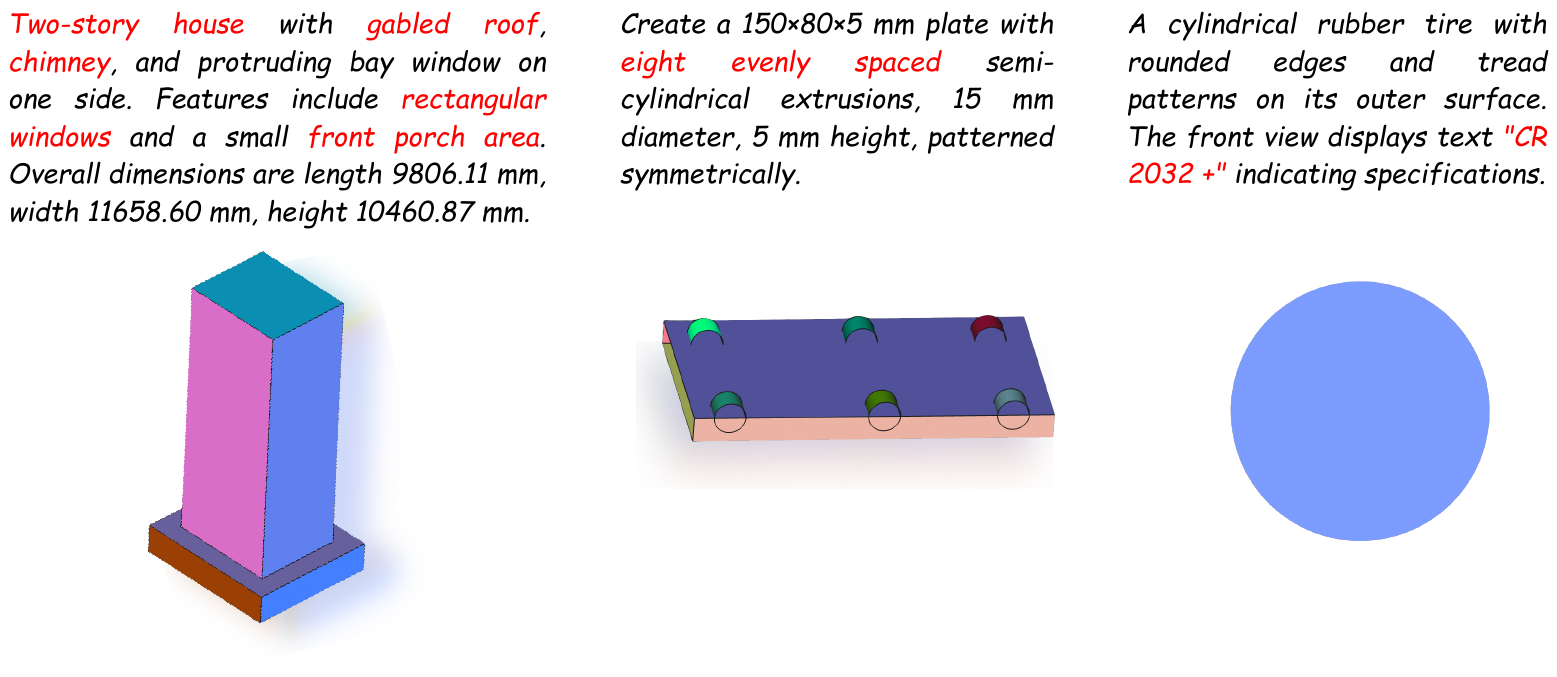}
    \caption{Failure cases of NURBGen illustrating limitations in handling complex prompts, geometric artifacts like self-intersections, and challenges in text engraving.}
    \label{fig:failure_case}
\end{figure}

\begin{table}[ht]
\def\arraystretch{1.2}%
\centering
\resizebox{\linewidth}{!}{
\begin{tabular}{lccccccc}
\hline
\textbf{Model} & \textbf{User(1k)}$\uparrow$ & \textbf{GPT$\uparrow$} & \textbf{IR$\downarrow$} & \textbf{CD$\downarrow$} & \textbf{HD$\downarrow$} & \textbf{JSD$\downarrow$} & \textbf{MMD$\downarrow$} \\
\hline
Undecided           &2.7 & 3.2 & --    & --     & --     & --     & --     \\
GPT-4o              & 1.5& 1.9 & 0.17 &7.2 &0.36 &72.87&4.17 \\
DeepCAD             & 5.6 &6.1 & 0.32& 10.28& 0.45& 89.77& 4.43\\
Text2CAD            &26.1 & 27.2 & 0.05& 9.66& 0.42& 85.27& 4.54\\
\textbf{NURBGen} &\textbf{64.1} & \textbf{61.6} &\textbf{0.018} & \textbf{4.43} &\textbf{0.25} & \textbf{57.94} & \textbf{2.14} \\
\hline
\end{tabular}
}
\caption{Quantitative comparison of text-to-CAD models. CD, JSD and MMD are multiplied by $10^2$.}
\label{tab:quantitative}
% \vspace*{-0.5\baselineskip}
\end{table}

\vspace{1mm}

\noindent \textbf{Results:} Table 1 shows the quantitative comparison of NURBGen with other baselines. Our model outperforms the prior baselines by a significant margin in both geometric and visual alignment. Notably, we achieve 60.8\% top-1 preference in human evaluation and 63.7\% in GPT-4o evaluation. Text2CAD ranks second, followed by DeepCAD and GPT-4o. Notably, NURBGen also achieves the lowest invalidity ratio (0.01), indicating strong geometric correctness in its output. In contrast, DeepCAD suffers from a higher invalidity rate (0.3), reflecting challenges in generating complete and consistent BRep geometry. As illustrated in Figure~\ref{fig:qualitative}, NURBGen generates CAD models that are better aligned with the input text, outperforming baselines in both fidelity and consistency.

\vspace{1mm}

\noindent \textbf{Caption Quality:} Since VLMs are prone to hallucinations~\cite{vlm_hallucinations}, we perform an evaluation to assess the accuracy of our automatically generated captions. We randomly sample 1,000 captions and provide them along with the corresponding six rendered views and associated CAD metadata to GPT-4o. GPT-4o is tasked with verifying the correctness of each caption. We observe an estimated accuracy of $\sim85\%$ across the sampled set. This indicates that our captioning pipeline produces high-quality and semantically meaningful descriptions.

% \vspace*{-0.5\baselineskip}
\section{Ablation Study}
\begin{figure}[t]
    \centering
    \includegraphics[width=0.96\linewidth]{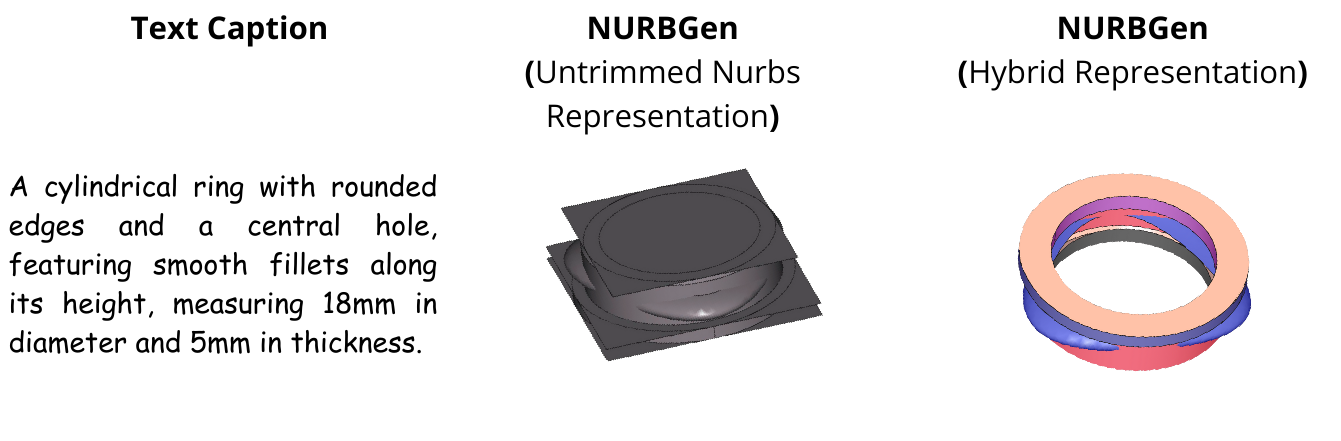}
        \caption{Comparison of NURBS-only (left) and hybrid (right) models, showing improved handling of thin and hole-adjacent regions.}
    \label{fig:ablation_study}
\end{figure}

% \vspace{1mm}
We conduct an ablation study to evaluate the effectiveness of our hybrid representation. Specifically, we fine-tune Qwen3-4B using only the untrimmed NURBS-based representation, without fallback to analytic primitives. We conduct human and GPT-4o evaluation using the same strategy mentioned before. We observe that the model trained with the hybrid representation scores 72\% by human and $79\%$ by GPT-4o. As shown in Figure~\ref{fig:ablation_study}, this leads to geometric artifacts and reconstruction errors, particularly near holes, sharp transitions, or regions where NURBS fitting is imprecise. This highlights the importance of our hybrid approach in achieving robust and accurate CAD reconstruction.

% \vspace*{-0.5\baselineskip}
\section{Limitation}
% \vspace{1mm}
\noindent Despite its strong performance, our approach has certain limitations. Figure~\ref{fig:failure_case} illustrates a few representative failure cases. For instance, in response to complex prompts (e.g., “\textit{Two-story house with gabled roof}...”), NURBGen struggles to capture fine-grained architectural structure. In rare cases, we also observe geometric artifacts such as self-intersections or topological inconsistencies, as seen in the second example. Additionally, NURBGen has difficulty reconstructing prompts with engraving text (third example).

% \vspace*{-0.5\baselineskip}
\section{Conclusion}
% \vspace{1mm}
We present \textbf{NURBGen}, the first framework for text-to-CAD generation using NURBS surfaces. NURBGen generates structured, editable NURBS representations from text prompts, which can be directly converted into B-Rep format using a fine-tuned LLM. To enable this, we generate \textbf{partABC}, a large-scale dataset of 300k part-level models from ABC with NURBS annotations and high-quality generated captions. We hope that this dataset will be a valuable resource for future research.  We further propose a hybrid representation that combines untrimmed NURBS with analytic primitives to address trimming artifacts while enhancing geometric robustness and token efficiency. Empirical results show that NURBGen surpasses existing state-of-the-art methods in geometric fidelity, as confirmed by expert evaluators. While the current model is constrained by a context window of $8192$, future work will explore long-context training and multimodal extensions to handle more complex assemblies. We believe that our work will position NURBS-based representations as a compelling alternative to design-history-based methods for future research in the evolving text-to-CAD domain.

\section{Acknowledgement}
This work was co-funded by the European Union under Horizon Europe, grant number 101135724, project LUMINOUS. However, the views and opinions expressed are those of the author(s) only and do not necessarily reflect those of the European Union. Neither the European Union nor the granting authority can be held responsible.
\bibliography{aaai2026}

@article{cad_additive_review,
	article-number = {7155},
	author = {Vido, Marcos and de Oliveira Neto, Geraldo Cardoso and Louren{\c c}o, Sergio Ricardo and Amorim, Marlene and Rodrigues, M{\'a}rio Jorge Ferreira},
	doi = {10.3390/app14167155},
	issn = {2076-3417},
	journal = {Applied Sciences},
	number = {16},
	title = {Computer-Aided Design and Additive Manufacturing for Automotive Prototypes: A Review},
	url = {https://www.mdpi.com/2076-3417/14/16/7155},
	volume = {14},
	year = {2024}}

@article{cad_status_engineering,
	author = {Wei Gao and Yunbo Zhang and Devarajan Ramanujan and Karthik Ramani and Yong Chen and Christopher B. Williams and Charlie C.L. Wang and Yung C. Shin and Song Zhang and Pablo D. Zavattieri},
	doi = {https://doi.org/10.1016/j.cad.2015.04.001},
	issn = {0010-4485},
	journal = {Computer-Aided Design},
	pages = {65-89},
	title = {The status, challenges, and future of additive manufacturing in engineering},
	url = {https://www.sciencedirect.com/science/article/pii/S0010448515000469},
	volume = {69},
	year = {2015}}

@Inproceedings{text2cad,
title={Text2CAD: Generating Sequential {CAD} Designs from Beginner-to-Expert Level Text Prompts},
author={Mohammad Sadil Khan and Sankalp Sinha and Sheikh Talha Uddin and Didier Stricker and Sk Aziz Ali and Muhammad Zeshan Afzal},
booktitle = {Advances in Neural Information Processing Systems},
pages = {7552--7579},
publisher = {Curran Associates, Inc.},
year={2024},
volume = {37},
url = {https://proceedings.neurips.cc/paper_files/paper/2024/file/0e5b96f97c1813bb75f6c28532c2ecc7-Paper-Conference.pdf},
}

@Inproceedings{cadsignet,
author = {Khan, Mohammad Sadil and Dupont, Elona and Ali, Sk Aziz and Cherenkova, Kseniya and Kacem, Anis and Aouada, Djamila},
title = {CAD-SIGNet: CAD Language Inference from Point Clouds using Layer-wise Sketch Instance Guided Attention},
booktitle = {Proceedings of the IEEE/CVF Conference on Computer Vision and Pattern Recognition (CVPR)},
month = {June},
year = {2024},
pages = {4713-4722}
}

@Inproceedings{cadllama,
      title={CAD-Llama: Leveraging Large Language Models for Computer-Aided Design Parametric 3D Model Generation}, 
      author={Jiahao Li and Weijian Ma and Xueyang Li and Yunzhong Lou and Guichun Zhou and Xiangdong Zhou},
      year={2025},
     booktitle = {Proceedings of the IEEE/CVF Conference on Computer Vision and Pattern Recognition (CVPR)},
month = {June},
      url={https://openaccess.thecvf.com/content/CVPR2025/papers/Li_CAD-Llama_Leveraging_Large_Language_Models_for_Computer-Aided_Design_Parametric_3D_CVPR_2025_paper.pdf}, 
}

@misc{point2cad,
      title={Point2CAD: Reverse Engineering CAD Models from 3D Point Clouds}, 
      author={Yujia Liu and Anton Obukhov and Jan Dirk Wegner and Konrad Schindler},
      year={2023},
      eprint={2312.04962},
      archivePrefix={arXiv},
      primaryClass={cs.CV},
      url={https://arxiv.org/abs/2312.04962}, 
}

@misc{img2cad,
      title={Img2CAD: Conditioned 3D CAD Model Generation from Single Image with Structured Visual Geometry}, 
      author={Tianrun Chen and Chunan Yu and Yuanqi Hu and Jing Li and Tao Xu and Runlong Cao and Lanyun Zhu and Ying Zang and Yong Zhang and Zejian Li and Linyun Sun},
      year={2024},
      eprint={2410.03417},
      archivePrefix={arXiv},
      primaryClass={cs.CV},
      url={https://arxiv.org/abs/2410.03417}, 
}

@misc{cadgpt,
      title={CADgpt: Harnessing Natural Language Processing for 3D Modelling to Enhance Computer-Aided Design Workflows}, 
      author={Timo Kapsalis},
      year={2024},
      eprint={2401.05476},
      archivePrefix={arXiv},
      primaryClass={cs.HC},
      url={https://arxiv.org/abs/2401.05476}, 
}

@article{cad-instruct,
	author = {Chaofan Lv and Jinsong Bao},
	doi = {https://doi.org/10.1016/j.cad.2025.103926},
	issn = {0010-4485},
	journal = {Computer-Aided Design},
	pages = {103926},
	title = {CADInstruct: A multimodal dataset for natural language-guided CAD program synthesis},
	url = {https://www.sciencedirect.com/science/article/pii/S0010448525000879},
	volume = {188},
	year = {2025}}

@misc{cadmium,
      title={CADmium: Fine-Tuning Code Language Models for Text-Driven Sequential CAD Design}, 
      author={Prashant Govindarajan and Davide Baldelli and Jay Pathak and Quentin Fournier and Sarath Chandar},
      year={2025},
      eprint={2507.09792},
      archivePrefix={arXiv},
      primaryClass={cs.GR},
      url={https://arxiv.org/abs/2507.09792}, 
}

@inproceedings{
text-to-cad,
title={Text-to-{CAD} Generation Through Infusing Visual Feedback in Large Language Models},
author={Ruiyu Wang and Yu Yuan and Shizhao Sun and Jiang Bian},
booktitle={Forty-second International Conference on Machine Learning},
year={2025},
url={https://openreview.net/forum?id=DW8oTCk2nF}
}

@InProceedings{deepcad,
    author    = {Wu, Rundi and Xiao, Chang and Zheng, Changxi},
    title     = {DeepCAD: A Deep Generative Network for Computer-Aided Design Models},
    booktitle = {Proceedings of the IEEE/CVF International Conference on Computer Vision (ICCV)},
    month     = {October},
    year      = {2021},
    pages     = {6772-6782}
}

@InProceedings{abc,
author = {Koch, Sebastian and Matveev, Albert and Jiang, Zhongshi and Williams, Francis and Artemov, Alexey and Burnaev, Evgeny and Alexa, Marc and Zorin, Denis and Panozzo, Daniele},
title = {ABC: A Big CAD Model Dataset For Geometric Deep Learning},
booktitle = {The IEEE Conference on Computer Vision and Pattern Recognition (CVPR)},
month = {June},
year = {2019}
}

@misc{neuronurbs,
      title={NeuroNURBS: Learning Efficient Surface Representations for 3D Solids}, 
      author={Jiajie Fan and Babak Gholami and Thomas Bäck and Hao Wang},
      year={2024},
      eprint={2411.10848},
      archivePrefix={arXiv},
      primaryClass={cs.CV},
      url={https://arxiv.org/abs/2411.10848}, 
}

@misc{qwen3,
      title={Qwen3 Technical Report}, 
      author={An Yang and Anfeng Li and Baosong Yang and Beichen Zhang and Binyuan Hui and Bo Zheng and et al},
      year={2025},
      eprint={2505.09388},
      archivePrefix={arXiv},
      primaryClass={cs.CL},
      url={https://arxiv.org/abs/2505.09388}, 
}

@article{nurbsdiff,
  author={Anjana Deva Prasad and Aditya Balu and Harshil Shah and Soumik Sarkar and Chinmay Hegde and Adarsh Krishnamurthy},
  title={NURBS-Diff: A Differentiable Programming Module for NURBS},
  year={2022},
  cdate={1640995200000},
  journal={Computer Aided Design},
  volume={146},
  pages={103199},
  url={https://doi.org/10.1016/j.cad.2022.103199}
}

@inproceedings{cadparser,
author = {Zhou, Shengdi and Tang, Tianyi and Zhou, Bin},
title = {CADParser: a learning approach of sequence modeling for B-Rep CAD},
year = {2023},
isbn = {978-1-956792-03-4},
url = {https://doi.org/10.24963/ijcai.2023/200},
doi = {10.24963/ijcai.2023/200},
booktitle = {Proceedings of the Thirty-Second International Joint Conference on Artificial Intelligence},
articleno = {200},
numpages = {9},
location = {Macao, P.R.China},
series = {IJCAI '23}
}

@misc{parsenet,
    title={ParSeNet: A Parametric Surface Fitting Network for 3D Point Clouds},
    author={Gopal Sharma and Difan Liu and Evangelos Kalogerakis and Subhransu Maji and Siddhartha Chaudhuri and Radomír Měch},
    year={2020},
    eprint={2003.12181},
    archivePrefix={arXiv},
    primaryClass={cs.CV}
}

@inproceedings{cadtranslator,
author = {Li, Xueyang and Song, Yu and Lou, Yunzhong and Zhou, Xiangdong},
title = {CAD Translator: An Effective Drive for Text to 3D Parametric Computer-Aided Design Generative Modeling},
year = {2024},
isbn = {9798400706868},
publisher = {Association for Computing Machinery},
address = {New York, NY, USA},
url = {https://doi.org/10.1145/3664647.3681549},
doi = {10.1145/3664647.3681549},
booktitle = {Proceedings of the 32nd ACM International Conference on Multimedia},
pages = {8461–8470},
numpages = {10},
location = {Melbourne VIC, Australia},
series = {MM '24}
}

@INPROCEEDINGS{cadopsnet,
  author={Dupont, Elona and Cherenkova, Kseniya and Kacem, Anis and Ali, Sk Aziz and Arzhannikov, Ilya and Gusev, Gleb and Aouada, Djamila},
  booktitle={2022 International Conference on 3D Vision (3DV)}, 
  title={CADOps-Net: Jointly Learning CAD Operation Types and Steps from Boundary-Representations}, 
  year={2022},
  volume={},
  number={},
  pages={114-123},
  doi={10.1109/3DV57658.2022.00024}}

@inproceedings{brepdetnet,
  title = {BRep Boundary and Junction Detection for CAD Reverse Engineering},
  author = {Ali, Sk Aziz and Khan, Mohammad Sadil and Stricker, Didier},
  booktitle = {IEEE International Conference on Computing and Machine Intelligence (ICMI)},
  year = {2024},
  organization = {IEEE},
}

@InProceedings{sharp2023,
    author    = {Mallis, Dimitrios and Aziz, Ali Sk and Dupont, Elona and Cherenkova, Kseniya and Karadeniz, Ahmet Serdar and Khan, Mohammad Sadil and Kacem, Anis and Gusev, Gleb and Aouada, Djamila},
    title     = {SHARP Challenge 2023: Solving CAD History and pArameters Recovery from Point Clouds and 3D Scans. Overview, Datasets, Metrics, and Baselines.},
    booktitle = {Proceedings of the IEEE/CVF International Conference on Computer Vision (ICCV) Workshops},
    month     = {October},
    year      = {2023},
    pages     = {1786-1795}
}

@InProceedings{caprinet,
    author    = {Yu, Fenggen and Chen, Zhiqin and Li, Manyi and Sanghi, Aditya and Shayani, Hooman and Mahdavi-Amiri, Ali and Zhang, Hao},
    title     = {CAPRI-Net: Learning Compact CAD Shapes With Adaptive Primitive Assembly},
    booktitle = {Proceedings of the IEEE/CVF Conference on Computer Vision and Pattern Recognition (CVPR)},
    month     = {June},
    year      = {2022},
    pages     = {11768-11778}
}

@misc{secadnet,
      title={SECAD-Net: Self-Supervised CAD Reconstruction by Learning Sketch-Extrude Operations}, 
      author={Pu Li and Jianwei Guo and Xiaopeng Zhang and Dong-ming Yan},
      year={2023},
      eprint={2303.10613},
      archivePrefix={arXiv},
      primaryClass={cs.CV},
      url={https://arxiv.org/abs/2303.10613}, 
}

@misc{pointnet,
      title={PointNet: Deep Learning on Point Sets for 3D Classification and Segmentation}, 
      author={Charles R. Qi and Hao Su and Kaichun Mo and Leonidas J. Guibas},
      year={2017},
      eprint={1612.00593},
      archivePrefix={arXiv},
      primaryClass={cs.CV},
      url={https://arxiv.org/abs/1612.00593}, 
}

@article{complexgen,
    author = {Haoxiang Guo and Shilin Liu and Hao Pan and Yang Liu and Xin Tong and Baining Guo},
    title = {ComplexGen: CAD Reconstruction by B-Rep Chain Complex Generation},
    year = {2022},
    issue_date = {July 2022},
    publisher = {Association for Computing Machinery},
    volume = {41},
    number = {4},
    url = {https://doi.org/10.1145/3528223.3530078},
    doi = {10.1145/3528223.3530078},
    journal = {ACM Trans. Graph. (SIGGRAPH)},
    month = jul,
    articleno = {129},
    numpages = {18}
}

@article{pointnet++,
      title={PointNet++: Deep Hierarchical Feature Learning on Point Sets in a Metric Space},
      author={Qi, Charles R and Yi, Li and Su, Hao and Guibas, Leonidas J},
      journal={arXiv preprint arXiv:1706.02413},
      year={2017}
    }

@inproceedings{brepnet,
    author    = {Lambourne, Joseph G. and Willis, Karl D.D. and Jayaraman, Pradeep Kumar and Sanghi, Aditya and Meltzer, Peter and Shayani, Hooman},
    title     = {BRepNet: A Topological Message Passing System for Solid Models},
    booktitle = {Proceedings of the IEEE/CVF Conference on Computer Vision and Pattern Recognition (CVPR)},
    month     = {June},
    year      = {2021},
    pages     = {12773-12782}
}

@article{brepgat,
	author = {Lee, Jinwon and Yeo, Changmo and Cheon, Sang-Uk and Park, Jun Hwan and Mun, Duhwan},
	doi = {10.1093/jcde/qwad106},
	eprint = {https://academic.oup.com/jcde/article-pdf/10/6/2384/56245545/qwad106.pdf},
	issn = {2288-5048},
	journal = {Journal of Computational Design and Engineering},
	month = {11},
	number = {6},
	pages = {2384-2400},
	title = {BRepGAT: Graph neural network to segment machining feature faces in a B-rep model},
	url = {https://doi.org/10.1093/jcde/qwad106},
	volume = {10},
	year = {2023}}

@misc{transcad,
      title={TransCAD: A Hierarchical Transformer for CAD Sequence Inference from Point Clouds}, 
      author={Elona Dupont and Kseniya Cherenkova and Dimitrios Mallis and Gleb Gusev and Anis Kacem and Djamila Aouada},
      year={2024},
      eprint={2407.12702},
      archivePrefix={arXiv},
      primaryClass={cs.CV},
      url={https://arxiv.org/abs/2407.12702}, 
}

@article{cadrille,
  title={cadrille: Multi-modal CAD Reconstruction with Online Reinforcement Learning},
  author={Maksim Kolodiazhnyi and Denis Tarasov and Dmitrii Zhemchuzhnikov and Alexander Nikulin and Ilya Zisman and Anna Vorontsova and Anton Konushin and Vladislav Kurenkov and Danila Rukhovich},
  journal={arXiv preprint arXiv:2505.22914},
  year={2025}
}

@misc{cad-mllm,
      title={CAD-MLLM: Unifying Multimodality-Conditioned CAD Generation With MLLM}, 
      author={Jingwei Xu and Zibo Zhao and Chenyu Wang and Wen Liu and Yi Ma and Shenghua Gao},
      year={2025},
      eprint={2411.04954},
      archivePrefix={arXiv},
      primaryClass={cs.CV},
      url={https://arxiv.org/abs/2411.04954}, 
}

@article{cadrecode,
  title={CAD-Recode: Reverse Engineering CAD Code from Point Clouds},
  author={Danila Rukhovich and Elona Dupont and Dimitrios Mallis and Kseniya Cherenkova and Anis Kacem and Djamila Aouada},
  journal={arXiv preprint arXiv:2412.14042},
  year={2024}
}

@article{fusion360,
    title={Fusion 360 Gallery: A Dataset and Environment for Programmatic CAD Construction from Human Design Sequences},
    author={Karl D. D. Willis and Yewen Pu and Jieliang Luo and Hang Chu and Tao Du and Joseph G. Lambourne and Armando Solar-Lezama and Wojciech Matusik},
    journal={ACM Transactions on Graphics (TOG)},
    volume={40},
    number={4},
    year={2021},
    publisher={ACM New York, NY, USA}
}

@misc{extrudenet,
      title={ExtrudeNet: Unsupervised Inverse Sketch-and-Extrude for Shape Parsing}, 
      author={Daxuan Ren and Jianmin Zheng and Jianfei Cai and Jiatong Li and Junzhe Zhang},
      year={2022},
      eprint={2209.15632},
      archivePrefix={arXiv},
      primaryClass={cs.CV},
      url={https://arxiv.org/abs/2209.15632}, 
}

@article{brepgen,
		  title={Brepgen: A b-rep generative diffusion model with structured latent geometry},
		  author={Xu, Xiang and Lambourne, Joseph and Jayaraman, Pradeep and Wang, Zhengqing and Willis, Karl and Furukawa, Yasutaka},
		  journal={ACM Transactions on Graphics (TOG)},
		  volume={43},
		  number={4},
		  pages={1--14},
		  year={2024},
		  publisher={ACM New York, NY, USA}
		}

@article{drpg,
  author = {Worchel, Markus and Alexa, Marc},
  title = {Differentiable Rendering of Parametric Geometry},
  journal = {ACM Transactions On Graphics.},
  year = {2023},
  month = {dec},
  volume = {42},
  number = {6},
  url = {https://doi.org/10.1145/3618387},
  doi = {10.1145/3618387},
}

@misc{llm_robotics,
      title={Large Language Models for Robotics: A Survey}, 
      author={Fanlong Zeng and Wensheng Gan and Yongheng Wang and Ning Liu and Philip S. Yu},
      year={2023},
      eprint={2311.07226},
      archivePrefix={arXiv},
      primaryClass={cs.RO},
      url={https://arxiv.org/abs/2311.07226}, 
}

@inproceedings{pointllm,
  title={PointLLM: Empowering Large Language Models to Understand Point Clouds},
  author={Xu, Runsen and Wang, Xiaolong and Wang, Tai and Chen, Yilun and Pang, Jiangmiao and Lin, Dahua},
  booktitle={European Conference on Computer Vision},
  year={2024}
}

@article{3dllm,
 author = {Hong, Yining and Zhen, Haoyu and Chen, Peihao and Zheng, Shuhong and Du, Yilun and Chen, Zhenfang and Gan, Chuang},
 title = {3D-LLM: Injecting the 3D World into Large Language Models},
 journal = {NeurIPS},
 year = {2023},
}

@article{llama_mesh,
    title={LLaMA-Mesh: Unifying 3D Mesh Generation with Language Models},
    author={Zhengyi Wang and Jonathan Lorraine and Yikai Wang and Hang Su and Jun Zhu and Sanja Fidler and Xiaohui Zeng},
    journal={arXiv preprint arXiv:2411.09595},
    year={2024}
}

@misc{llama3,
      title={The Llama 3 Herd of Models}, 
      author={Aaron Grattafiori and et al},
      year={2024},
      eprint={2407.21783},
      archivePrefix={arXiv},
      primaryClass={cs.AI},
      url={https://arxiv.org/abs/2407.21783}, 
}

@misc{llm_cad_survey,
      title={Large Language Models for Computer-Aided Design: A Survey}, 
      author={Licheng Zhang and Bach Le and Naveed Akhtar and Siew-Kei Lam and Tuan Ngo},
      year={2025},
      eprint={2505.08137},
      archivePrefix={arXiv},
      primaryClass={cs.LG},
      url={https://arxiv.org/abs/2505.08137}, 
}

@InProceedings{marvel,
    author    = {Sinha, Sankalp and Khan, Mohammad Sadil and Usama, Muhammad and Sam, Shino and Stricker, Didier and Ali, Sk Aziz and Afzal, Muhammad Zeshan},
    title     = {MARVEL-40M+: Multi-Level Visual Elaboration for High-Fidelity Text-to-3D Content Creation},
    booktitle = {Proceedings of the Computer Vision and Pattern Recognition Conference (CVPR)},
    month     = {June},
    year      = {2025},
    pages     = {8105-8116}
}

@misc{internvl3,
      title={InternVL3: Exploring Advanced Training and Test-Time Recipes for Open-Source Multimodal Models}, 
      author={Jinguo Zhu and Weiyun Wang and Zhe Chen and Zhaoyang Liu and Shenglong Ye and Lixin Gu and Hao Tian and Yuchen Duan and Weijie Su and Jie Shao and Zhangwei Gao and Erfei Cui and Xuehui Wang and Yue Cao and Yangzhou Liu and Xingguang Wei and Hongjie Zhang and Haomin Wang and Weiye Xu and Hao Li and Jiahao Wang and Nianchen Deng and Songze Li and Yinan He and Tan Jiang and Jiapeng Luo and Yi Wang and Conghui He and Botian Shi and Xingcheng Zhang and Wenqi Shao and Junjun He and Yingtong Xiong and Wenwen Qu and Peng Sun and Penglong Jiao and Han Lv and Lijun Wu and Kaipeng Zhang and Huipeng Deng and Jiaye Ge and Kai Chen and Limin Wang and Min Dou and Lewei Lu and Xizhou Zhu and Tong Lu and Dahua Lin and Yu Qiao and Jifeng Dai and Wenhai Wang},
      year={2025},
      eprint={2504.10479},
      archivePrefix={arXiv},
      primaryClass={cs.CV},
      url={https://arxiv.org/abs/2504.10479}, 
}

@inproceedings{
adamw,
title={Decoupled Weight Decay Regularization},
author={Ilya Loshchilov and Frank Hutter},
booktitle={International Conference on Learning Representations},
year={2019},
url={https://openreview.net/forum?id=Bkg6RiCqY7},
}

@misc{lora,
      title={LoRA: Low-Rank Adaptation of Large Language Models}, 
      author={Edward J. Hu and Yelong Shen and Phillip Wallis and Zeyuan Allen-Zhu and Yuanzhi Li and Shean Wang and Lu Wang and Weizhu Chen},
      year={2021},
      eprint={2106.09685},
      archivePrefix={arXiv},
      primaryClass={cs.CL},
      url={https://arxiv.org/abs/2106.09685}, 
}

@misc{vlm_hallucinations,
      title={A Survey on Hallucination in Large Vision-Language Models}, 
      author={Hanchao Liu and Wenyuan Xue and Yifei Chen and Dapeng Chen and Xiutian Zhao and Ke Wang and Liping Hou and Rongjun Li and Wei Peng},
      year={2024},
      eprint={2402.00253},
      archivePrefix={arXiv},
      primaryClass={cs.CV},
      url={https://arxiv.org/abs/2402.00253}, 
}

@article{nurbot,
	author = {Yang, Shaoliang and Wang, Jun and Wang, Kang},
	doi = {10.1115/1.4066549},
	eprint = {https://asmedigitalcollection.asme.org/mechanicaldesign/article-pdf/147/3/031703/7389365/md\_147\_3\_031703.pdf},
	issn = {1050-0472},
	journal = {Journal of Mechanical Design},
	month = {10},
	number = {3},
	pages = {031703},
	title = {NURBS-OT: An Advanced Model for Generative Curve Modeling},
	url = {https://doi.org/10.1115/1.4066549},
	volume = {147},
	year = {2024}}

@article{nurb_optimization,
	author = {Orest Mykhaskiv and Mladen Banovi{\'c} and Salvatore Auriemma and Pavanakumar Mohanamuraly and Andrea Walther and Herve Legrand and Jens-Dominik M{\"u}ller},
	doi = {10.1080/16864360.2018.1462881},
	eprint = {https://doi.org/10.1080/16864360.2018.1462881},
	journal = {Computer-Aided Design and Applications},
	number = {6},
	pages = {916--926},
	publisher = {Taylor \& Francis},
	title = {NURBS-based and parametric-based shape optimization with differentiated CAD kernel},
	url = {https://doi.org/10.1080/16864360.2018.1462881},
	volume = {15},
	year = {2018}}

@article{nurb_survey,
	author = {Wolfgang B{\"o}hm and Gerald Farin and J{\"u}rgen Kahmann},
	doi = {https://doi.org/10.1016/0167-8396(84)90003-7},
	issn = {0167-8396},
	journal = {Computer Aided Geometric Design},
	number = {1},
	pages = {1-60},
	title = {A survey of curve and surface methods in CAGD},
	url = {https://www.sciencedirect.com/science/article/pii/0167839684900037},
	volume = {1},
	year = {1984}}

@misc{cibulkajoint,
  title={Joint Mathematical Conference CSASC 2010},
  author={Cibulka, Josef and Lidick{\`y}, Bernard and Tesa{\v{r}}, Marek},
}

% \input{ReproducibilityChecklist/LaTeX/ReproducibilityChecklist}

% FOR ARXIV VERSION
\twocolumn[{
     \begin{center}
         {\huge\bfseries Appendix}
     \end{center}
     \vspace{1em}
 }]
% FOR ARXIV VERSION

 \section{CAD Representation}
\begin{figure*}
    \centering
    \includegraphics[width=1\linewidth]{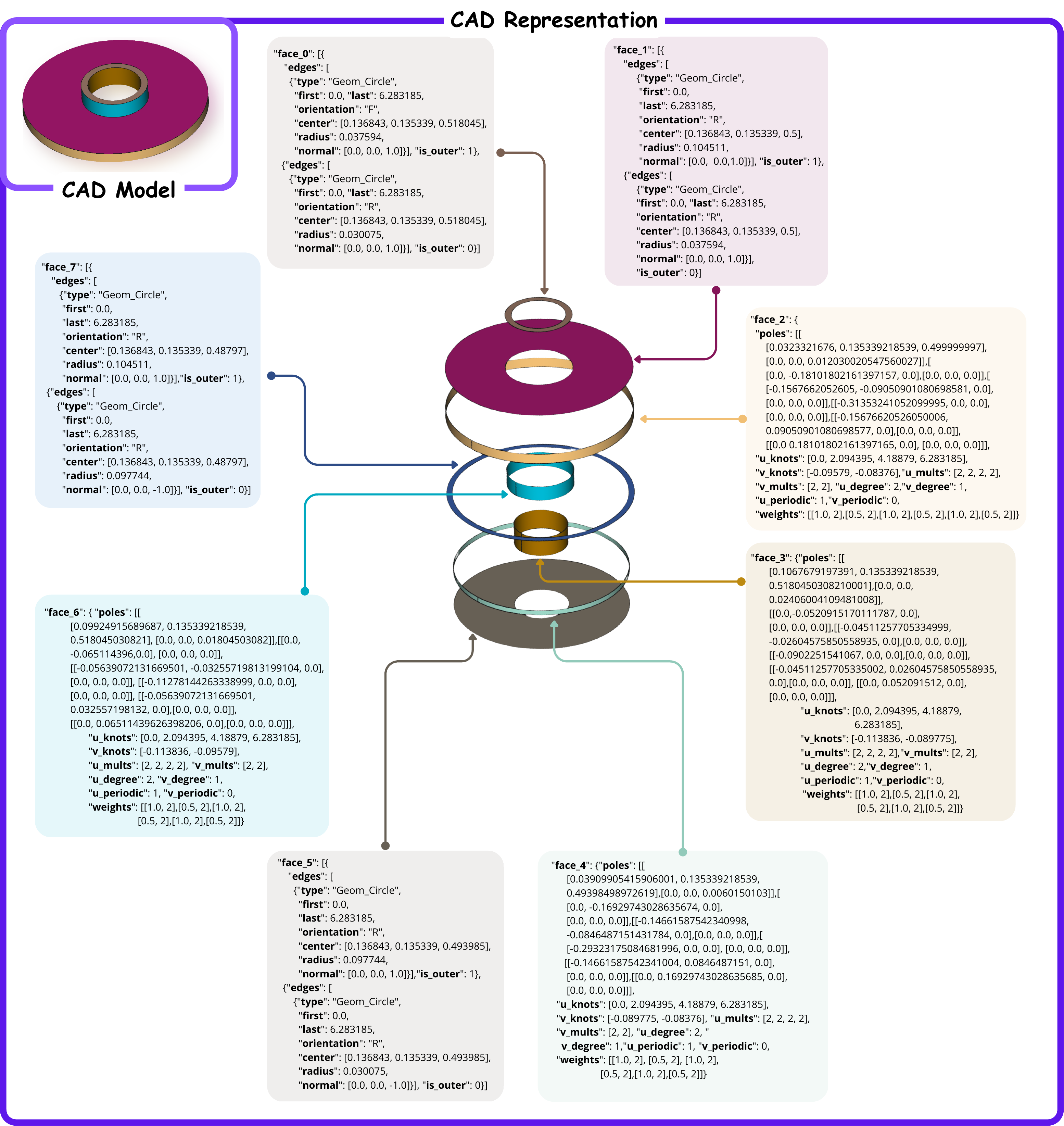}
    \caption{Our proposed hybrid CAD representation.}
    \label{fig:cad_representation_1}
\end{figure*}

We represent a BRep solid using a sequence of faces. Each face is either a NURB surface or contains analytical primitives (lines, circle, bsplines, and so on). Below, we provide how to parameterize them in our CAD representation. Figure~\ref{fig:cad_representation_1} showcases an example CAD representation.
\vspace{2mm}

\noindent \textbf{1. NURBS}:
\begin{itemize}
\item \texttt{Poles}: Control points, represented as a 2D array of 3D points defining the control net of the surface.
\item \texttt{Weights}: A 2D array of real numbers associated with the poles. Defines the rational nature of the surface. Optional for non-rational surfaces.
\item \texttt{u\_knots}, \texttt{v\_knots}: Non-decreasing sequences of real numbers defining the knot vectors in the $u$ and $v$ parametric directions.
\item \texttt{u\_mults}, \texttt{v\_mults}: Integer sequences representing the multiplicity of each knot in the $u$ and $v$ directions, respectively.
\item \texttt{u\_degree}, \texttt{v\_degree}: Degree of the B-spline basis functions in the $u$ and $v$ directions.
\item \texttt{u\_periodic}, \texttt{v\_periodic}: Boolean flags indicating whether the surface is periodic in each parametric direction.
\end{itemize}
\noindent \textbf{2. Line}

\begin{itemize}
    \item \texttt{start}: \((x, y, z)\) coordinates of the starting point.
    \item \texttt{end}: \((x, y, z)\) coordinates of the ending point.
\end{itemize}

\noindent \textbf{3. Circle}

\begin{itemize}
    \item \texttt{center}: \((x, y, z)\) coordinates of the circle’s center.
    \item \texttt{normal}: \((x, y, z)\) direction vector normal to the plane of the circle.
    \item \texttt{radius}: Radius of the circle.
    \item \texttt{first}, \texttt{last}: Start and end angles (in radians) defining an arc on the circle. For a semicircle, \texttt{first = 0.0}, \texttt{last = \(\pi\)}.
\end{itemize}

\noindent \textbf{4. Ellipse}

\begin{itemize}
    \item \texttt{center}: \((x, y, z)\) coordinates of the ellipse’s center.
    \item \texttt{normal}: \((x, y, z)\) direction vector normal to the ellipse’s plane.
    \item \texttt{major\_radius}: Length of the major axis.
    \item \texttt{minor\_radius}: Length of the minor axis.
    \item \texttt{first}, \texttt{last}: Start and end angles (in radians) defining an arc on the ellipse. For a full ellipse, \texttt{first = 0.0}, \texttt{last = \(2\pi\)}.
\end{itemize}

\noindent \textbf{5. Bezier Curve}

\begin{itemize}
    \item \texttt{poles}: List of control points \((x, y, z)\).
    \item \texttt{degree}: Degree of the Bezier curve.
    \item \texttt{first}, \texttt{last}: Parametric domain range.
\end{itemize}

\noindent \textbf{6. B-spline Curve}

\begin{itemize}
    \item \texttt{poles}: List of control points \((x, y, z)\).
    \item \texttt{degree}: Degree of the B-spline curve.
    \item \texttt{knots}: Knot vector (non-decreasing real numbers).
    \item \texttt{mults}: Corresponding multiplicities of knots.
    \item \texttt{weights} (optional): Weights for rational B-splines (if omitted, assumed to be 1.0).
    \item \texttt{is\_periodic}: Boolean flag indicating if the curve is periodic.
    \item \texttt{first}, \texttt{last}: Parametric range.
\end{itemize}

\begin{figure*}
    \centering
    \includegraphics[width=0.92\linewidth]{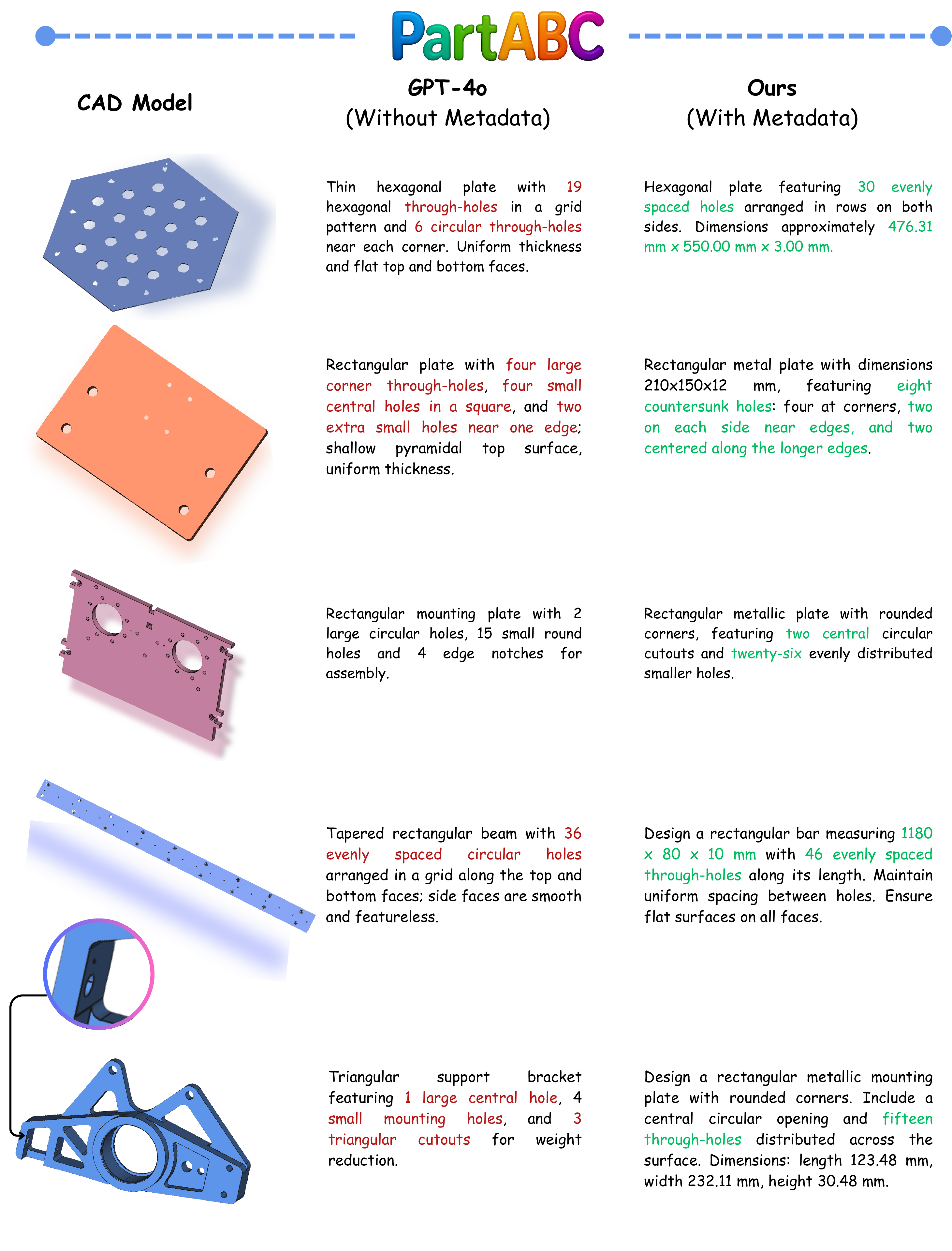}
    \caption{\textbf{Impact of metadata-guided annotation in our annotation pipeline}. From left to right: CAD model, caption generated by GPT-4o without metadata, and caption from our pipeline using metadata such as dimensions, and hole count.}
    \label{fig:metadata_impact_1}
\end{figure*}

\begin{figure*}
    \centering
    \includegraphics[width=0.95\linewidth]{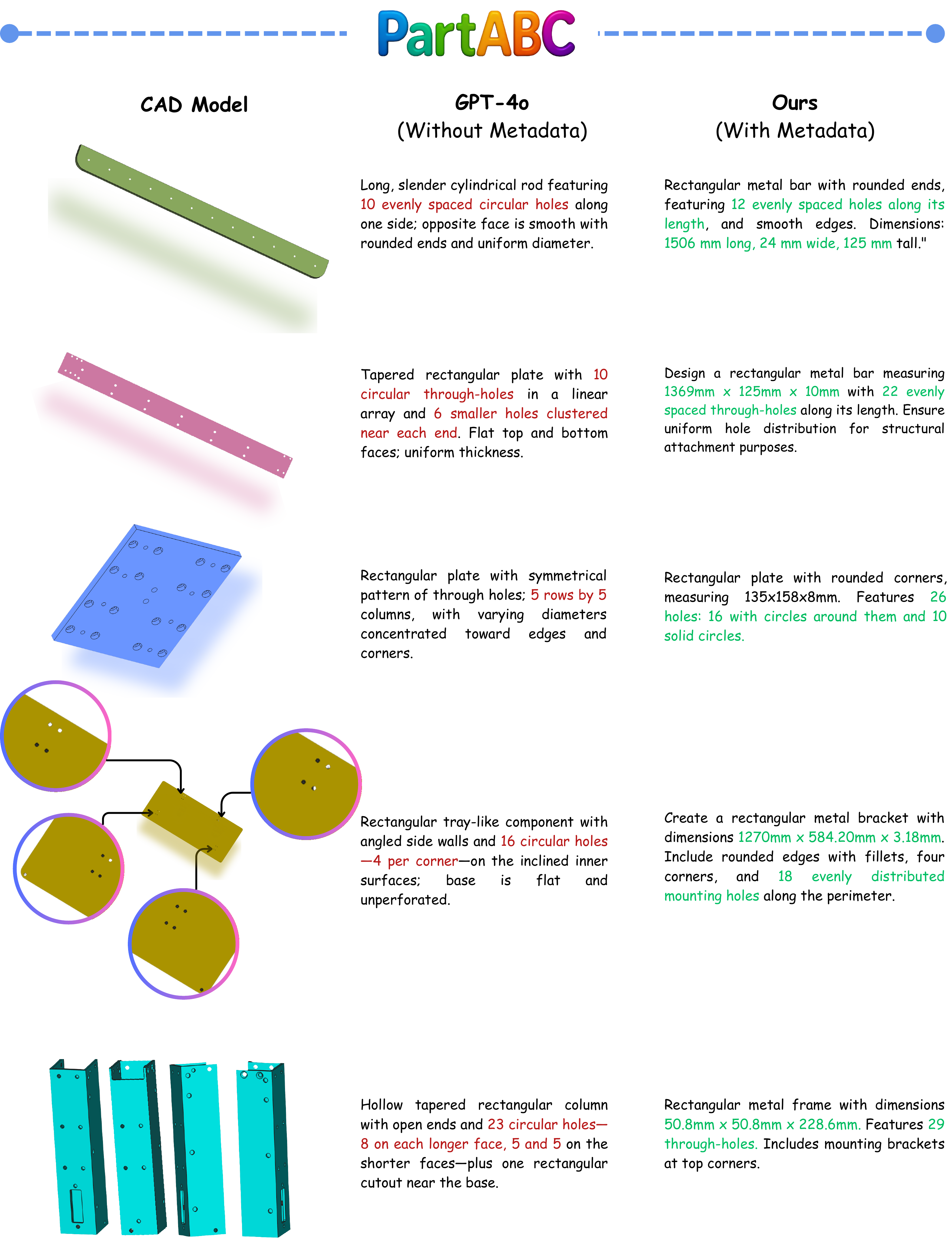}
    \caption{\textbf{Impact of metadata-guided annotation in our annotation pipeline}. From left to right: CAD model, caption generated by GPT-4o without metadata, and caption from our pipeline using metadata such as dimensions, and hole count.}
    \label{fig:metadata_impact_2}
\end{figure*}

\section{Impact of Metadata on Caption Quality}

Figure~\ref{fig:metadata_impact_1} and~\ref{fig:metadata_impact_2} highlight the crucial role of metadata in driving accurate and informative captions. With access to structural cues such as \textit{dimensions} and \textit{hole count}, InternVL3-$13$B produces descriptions that are both precise and grounded in geometry. In contrast, GPT-4o, when prompted without metadata, often overlooks or misstates these critical features.

% This underscores the importance of metadata-guided annotation in enabling fine-grained, engineering-relevant language grounding.

\section{More Qualitative Results}
\begin{figure*}[ht]
    \centering
    \includegraphics[width=0.8\linewidth]{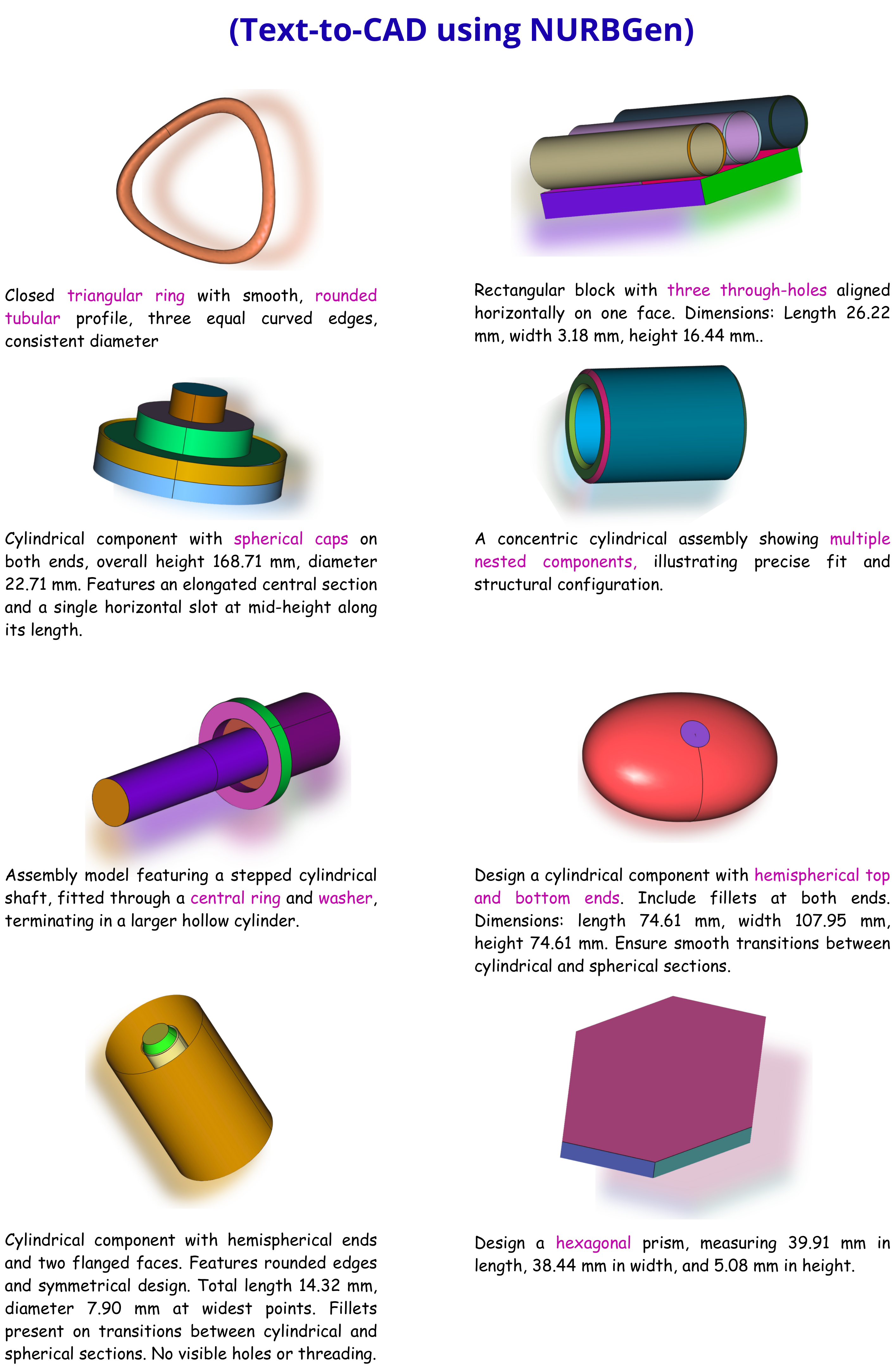}
    \caption{Text-to-CAD generation using NURBGen.}
    \label{fig:qual_1}
\end{figure*}

\begin{figure*}[t]
    \centering
    \includegraphics[width=0.8\linewidth]{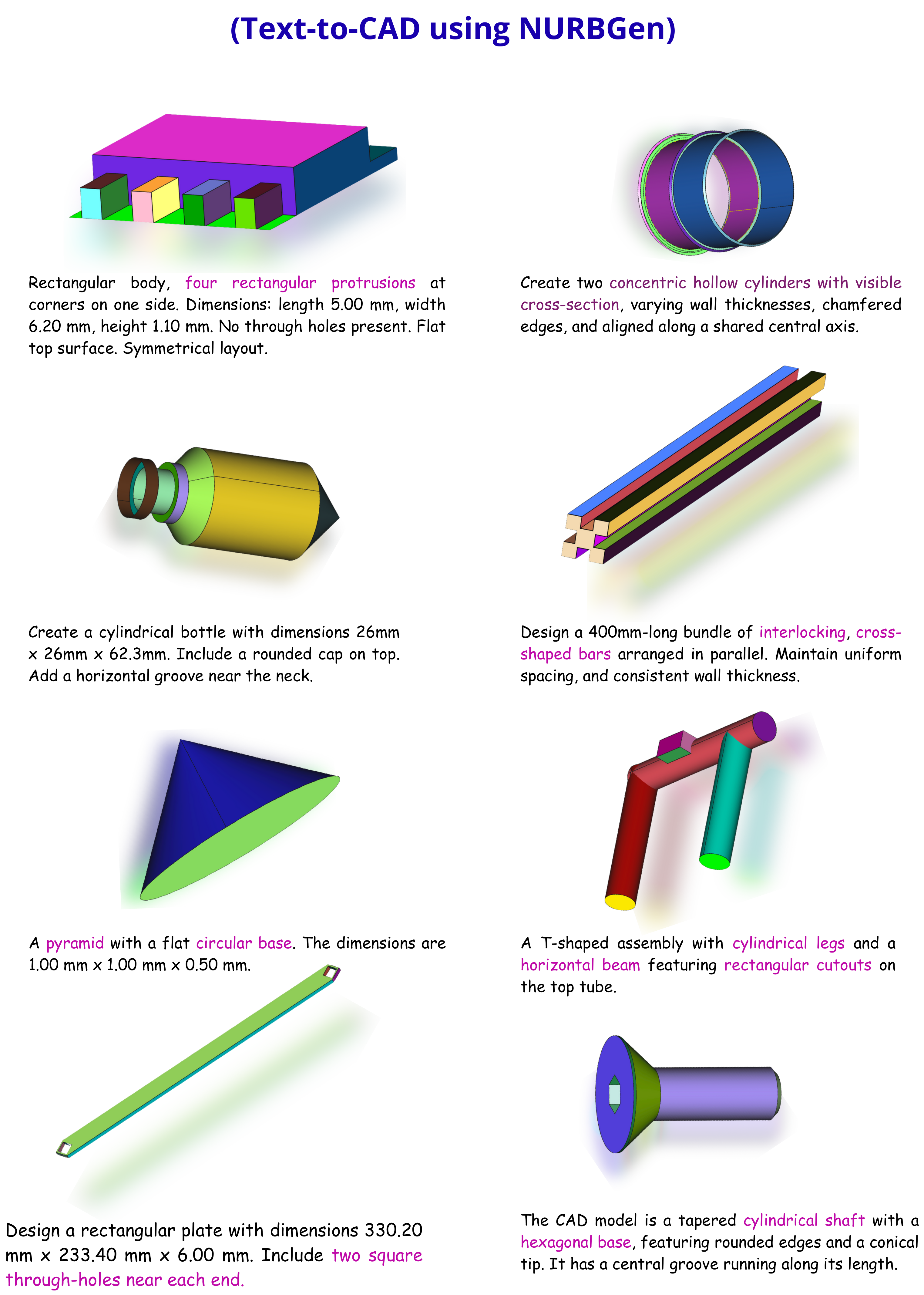}
    \caption{Text-to-CAD generation using NURBGen.}
    \label{fig:qual_2}
\end{figure*}

Figure~\ref{fig:qual_1} and Figure~\ref{fig:qual_2} showcase additional results from our text-to-CAD generation using NURBGen. Notably, examples such as (Column 1, Row 1) and (Column 2, Row 3) in Figure~\ref{fig:qual_1}, and (Column 1, Row 3) in Figure~\ref{fig:qual_2}, show shapes that are currently infeasible to generate using existing design-history-based text-to-CAD approaches due to lack of CAD operations such as loft, sweep and revolution in the training datasets.

\end{document}